\documentclass[10pt,twocolumn,letterpaper]{article}

\usepackage{iccv}
\usepackage{times}
\usepackage{epsfig}
\usepackage{graphicx}
\usepackage{amsmath}
\usepackage{amssymb}

\usepackage{amsthm}
\newtheorem{theorem}{Theorem}

\newtheorem{definition}{Definition}
\usepackage{multirow}
\usepackage{float}
\usepackage{longtable}
\usepackage{colortbl}
\usepackage{wrapfig}
\usepackage{dsfont}
\usepackage{amssymb}
\usepackage{bbding}
\usepackage{booktabs}
\usepackage{float}
\definecolor{Gray}{gray}{0.9}
\usepackage{soul}
\usepackage[accsupp]{axessibility} % I
\usepackage[breaklinks=true,bookmarks=false]{hyperref}

\iccvfinalcopy % *** Uncomment this line for the final submission

\def\iccvPaperID{7219} % *** Enter the ICCV Paper ID here

\ificcvfinal\fi

\begin{document}

%%%%%%%%% TITLE
\title{Understanding Self-attention Mechanism via Dynamical System Perspective}

\author{Zhongzhan Huang$^1$\thanks{co-first author \quad $\dagger$ corresponding author} \quad Mingfu Liang$^{2*}$ \quad Jinghui Qin$^3$ \quad Shanshan Zhong$^1$ \quad Liang Lin$^{1\dagger}$\\
$^1$Sun Yat-sen University \quad $^2$Northwestern University, USA \quad $^3$Guangdong University of Technology\\
{\tt\small huangzhzh23,zhongshsh5@mail2.sysu.edu.cn, mingfuliang2020@u.northwestern.edu,}\\
{\tt\small  scape1989@gmail.com, linliang@ieee.org}
}
\maketitle
\ificcvfinal\thispagestyle{empty}\fi

%%%%%%%%% ABSTRACT
\begin{abstract}
The self-attention mechanism (SAM) is widely used in various fields of artificial intelligence and has successfully boosted the performance of different models. However, current explanations of this mechanism are mainly based on intuitions and experiences, while there still lacks direct modeling for how the SAM helps performance. To mitigate this issue, in this paper, based on the dynamical system perspective of the residual neural network, we first show that the intrinsic stiffness phenomenon (SP) in the high-precision solution of ordinary differential equations (ODEs) also widely exists in high-performance neural networks (NN). Thus the ability of NN to measure SP at the feature level is necessary to obtain high performance and is an important factor in the difficulty of training NN. Similar to the adaptive step-size method which is effective in solving stiff ODEs, we show that the SAM is also a stiffness-aware step size adaptor that can enhance the model's representational ability to measure intrinsic SP by refining the estimation of stiffness information and generating adaptive attention values, which provides a new understanding about why and how the SAM can benefit the model performance. This novel perspective can also explain the lottery ticket hypothesis in SAM, design new quantitative metrics of representational ability, and inspire a new theoretic-inspired approach, StepNet. Extensive experiments on several popular benchmarks demonstrate that StepNet can extract fine-grained stiffness information and measure SP accurately, leading to significant improvements in various visual tasks.

\end{abstract}

\begin{figure}[t]
    \centering
    \includegraphics[width=1.0\linewidth]{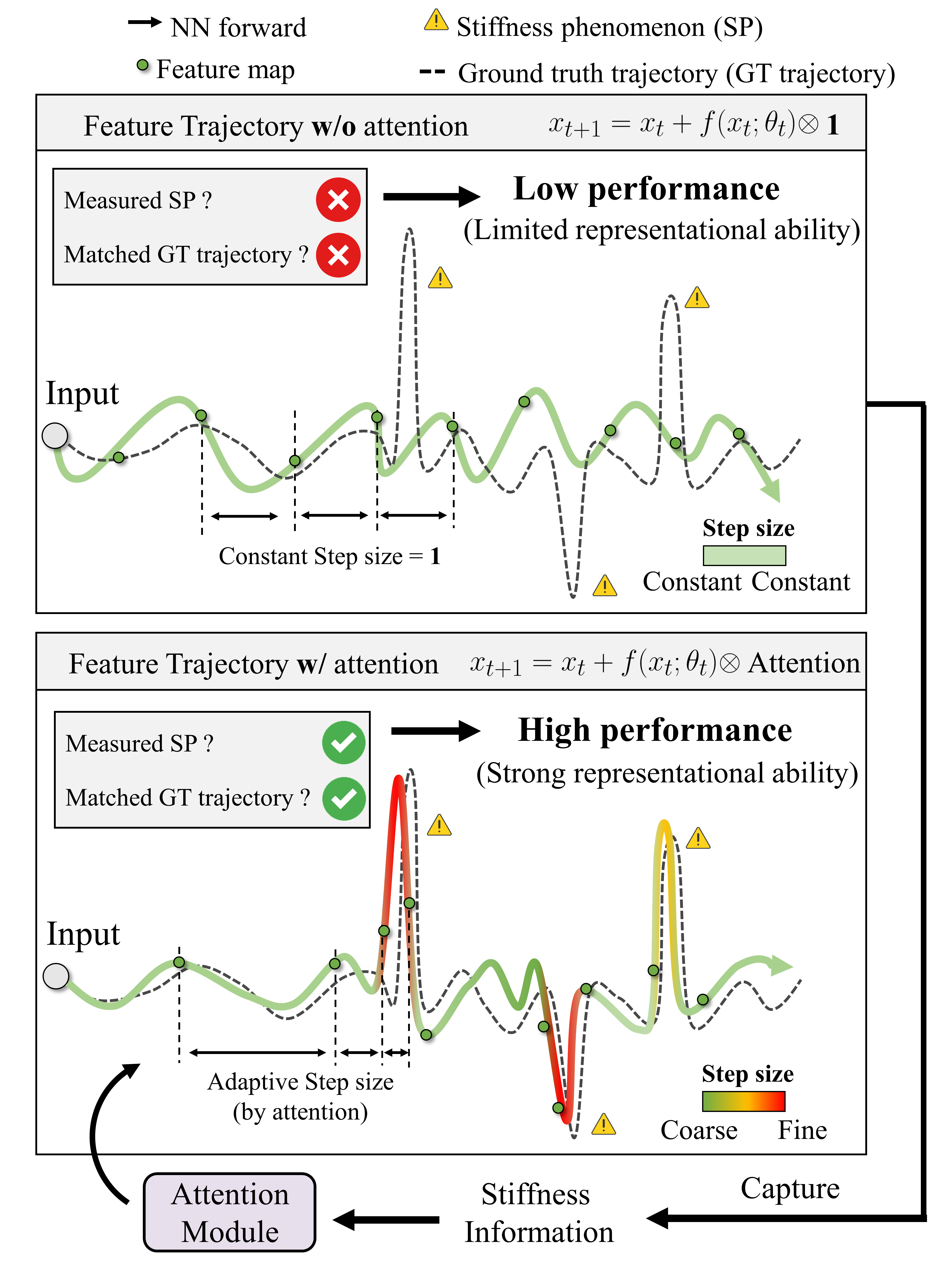}
    \vspace{-15pt}
    \caption{The feature trajectory w/ and w/o attention. The GT trajectory, generated by the best-performing model, possesses the inherent property of SP. The SAM, acting as a stiffness-aware step size adaptor, effectively extracts stiffness information and has a strong representational ability to measure the inherent SP to closely approach a GT trajectory, leading to high performance.  } 
    \vspace{-0.2cm}
    \label{fig:net}
    \vspace{-0.3cm}
\end{figure}
%  	 	\begin{figure*}[t]
% 	 		\centering
% 	 		\includegraphics[width=1.0\linewidth]{new.png}
% 	 				\vspace{-0.1cm}
% 	 		\caption{The feature trajectory with or without attention. The self-attention mechanism is
% a stiffness-aware step size adaptor that can refine the estimation of stiffness information and effectively alleviate the error accumulation due to the stiffness issue in ground truth trajectory. } 
% 	 \vspace{-0.2cm}
% 	 		\label{fig:net}
% 	 		\vspace{-0.3cm}
% 	 	\end{figure*}

%%%%%%%%% BODY TEXT
\section{Introduction}
\label{sec:intro}

The self-attention mechanism~(SAM) \cite{liu2021swin,guo2022attention,han2022survey,chen2021visformer,mao2022towards,chen2021crossvit} is widely used in various artificial intelligence fields and has successfully improved the models' performance in a number of vision tasks, including image classification \cite{hu2018squeeze,woo2018cbam,park2018bam}, object detection \cite{li2019spatial,wang2020eca,huang2022layer}, instance segmentation \cite{chen2020supervised,ren2017end}, image super-resolution \cite{zhang2018image,niu2020single,qin2020multi}, etc. However, most previous works lay emphasis on designing a new self-attention method, and intuitively or heuristically exploring how the self-attention mechanism helps the performance. For example, many popular channel attention methods~\cite{hu2018squeeze,park2018bam,wang2020eca,lee2019srm} consider the attention values as the soft weight of the channels, leading to the importance reassignment of feature maps. These soft weights can also be seen as a gate mechanism ~\cite{woo2018cbam,huang2020dianet} to control the forward transmission of information flow, which are usually applied to neural network pruning and neural architecture search \cite{liu2018darts,zhang2021rs}. Another viewpoint~\cite{liang2020instance} argues that the self-attention mechanism can help to regulate the noise by enhancing instance-specific information to obtain a better regularization effect. Moreover, the receptive field \cite{yang2018attention,zhang2021efficient,zhou2022efficient} and long-range dependency \cite{zhu2021lightweight,hao2020new,wang2020cluster} are also used to understand the role of self-attention. Although these explanations describe the behavior of self-attention mechanisms to some extent, the relationship between the SAM and model performance is still ambiguous. 

To establish a more specific modeling between the SAM and model performance, 
in this paper, we rethink the role of the SAM with the dynamical systems perspective of neural networks~(NN) with residual blocks. Specifically, we first define the stiffness phenomenon (SP) and ground truth (GT)  trajectory of NNs at the feature level based on stiffness metrics and the ground truth solution of ODEs. 

Next, we find that the intrinsic SP observed in the high-precision solutions of ODEs is also prevalent in high-performance NNs. This observation implies that the representational ability of NN to measure SP at the feature level is necessary to obtain high performance, as shown in Fig \ref{fig:net}, which is hindering the learning of neural networks, and advanced training strategies are needed to achieve this requirement. 

Similar to the adaptive step-size method which is effective in solving stiff ODEs, we theoretically and empirically demonstrate that the SAM is a stiffness-aware step size adaptor that can refine the estimation of stiffness information and generate the adaptive attention values to measure intrinsic SP for approaching a GT trajectory which has the upper bound performance, leading to high accuracy. This novel perspective can also explain the lottery ticket hypothesis in SAM (LTH4SA) \cite{huang2022lottery}, design new quantitative metrics of representational ability, and inspire a new theoretic-inspired approach, StepNet. Extensive experiments on several popular benchmarks show the effectiveness of StepNet in various vision tasks, including image classification and object detection. Our contributions are summarized as follows: 
\begin{enumerate}%[label=(\roman*)]
    \item We propose a novel understanding of the SAM and reveal a close connection between the SAMs and the numerical solution of stiff ODEs, which is an effective explanation for understanding why and how the SAM enhances the performance of NNs.
    \item  Based on our novel views of SAMs, we explain the lottery ticket hypothesis in SAM, design new quantitative metrics of representational ability, and propose a powerful theoretic-inspired approach, StepNet.
    %彩票假设，一致的
\end{enumerate}

% 结合图1
% \section{Related Work}
% \subsection{Self-attention Mechanism}
% \subsection{The Dynamical System Perspective}

\section{The Stiffness and Self-attention Mechanism}
\label{sec:theory}

In this section, we first introduce the concepts of stiffness in ODEs, SAM, and the dynamical system perspective for the NN with residual blocks. Then we further explore the SP in NNs and connect it with the SAM, which finally motivates us to propose a theoretic inspire approach.

\subsection{Preliminaries and Related Works}
\label{sec:preli}
\subsubsection{The Dynamical System Perspective of NN}
\label{sec:view}

There are many well-known network architectures that have the residual blocks, like ResNet \cite{he2016deep}, UNet \cite{huang2020unet}, Transformer \cite{liu2021swin}, ResNeXt \cite{xie2017aggregated}, etc. The residual blocks in one stage can be written as 
\vspace{-5pt}
\begin{equation}
    x_{t+1} = x_t + f(x_t;\theta_t),
    \label{eq:resnet}
\end{equation}
where $x_t \in \mathbb{R}^d$ is the input of NN $f(\cdot;\theta_t)$ with the learnable parameters $\theta_t$ in $t^{\rm th}$ block. Many recent works \cite{chang2017multi,DBLP:conf/icml/LuZLD18,chen2018neural,weinan2017proposal,queiruga2020continuous,zhu2022convolutional,meunier2022dynamical} have established an insightful connection between residual blocks and dynamical systems, which reveal that the residual blocks can be interpreted as one step of a forward numerical method, \textit{i.e.}, $\mathbf{u}_{t+1} = \mathbf{u}_t + S(\mathbf{u}_t; \mathbf{f},\Delta t)\cdot \Delta t$, for the numerical solution of an ODE as Eq.(\ref{eqn:ode}):
\vspace{-5pt}
\begin{align}
%\frac{\text{d}\mathbf{u}}{\text{d}t}
\text{d}\mathbf{u}(t)/\text{d}t = \mathbf{f}[\mathbf{u}(t)], \quad \mathbf{u}(0) = \mathbf{c}_0,
\label{eqn:ode}
\end{align}
where $\mathbf{c}_0$ represents an initial condition, which corresponds to the input of the residual network. $\mathbf{u}(t) \equiv \mathbf{u}_t$ is a time-dependent $d$-dimensional state, which is used to describe the input feature $x_t$ in $t^{\rm th}$ block. The output of neural network $f(\cdot;\theta_t)$ in $t^{\rm th}$ block can be regarded as an integration $S(\mathbf{u}_t;\mathbf{f},\Delta t)$ with step size $\Delta t$ using a numerical method $S$, e.g., the Forward Euler method \cite{shampine2018numerical}.% $S(\mathbf{u}_t; \mathbf{f},\Delta t)\cdot \Delta t\equiv \mathbf{f}(\mathbf{u}_t)\Delta t$. 

% \begin{table}[htbp]
%   \centering
%   \caption{The dynamical system perspective for residual neural networks.}
%   \vspace{-0.1cm}
%     \begin{tabular}{ll}
%     \toprule
%     \multicolumn{1}{c}{\textbf{Dynamical system}} & \multicolumn{1}{c}{\textbf{ResNet}} \\
%     \midrule
%     $\mathbf{u}_{t+1} = \mathbf{u}_t + S(\mathbf{u}_t; \mathbf{f},\Delta t)$     & $x_{t+1} = x_t + f(x_t;\theta_t)$  \\
%     \bottomrule
%     \end{tabular}%
%   \label{tab:dyandres}%
% \end{table}%

\subsubsection{The Stiffness in ODEs}
\label{sec:stiff_math}
In mathematics, a stiff equation is a differential equation \cite{lambert1976initial}, like Eq.(\ref{eqn:ode}), for which the numerical methods for solving that equation are numerically unstable, leading to poor prediction. For most ODEs, the stiffness is universal and intrinsic \cite{lambert1991numerical}.
When the solution is unstable, we can use a fine step size $\Delta t$ instead of a coarse step size to obtain finer differentiation, resulting in high-precision integration. Therefore, utilizing an adaptive step size based on a specific numerical method is the most straightforward way to solve stiffness ODEs, like the improvement from Forth order Runge–Kutta method \cite{butcher2016numerical} to Runge–Kutta–Fehlberg method \cite{watts1975runge,fehlberg1969low}. 
However, there is no universally accepted mathematical definition of stiffness~\cite{lambert1991numerical}, but the main idea is that the equation includes some terms that can lead to rapid variation in the solution. 

Therefore, to quantify the stiffness to a certain extent, some simplified indexes are proposed, like the versatile stiffness index (SI) \cite{aiken1985stiff} $\zeta_{\text{SI}}$ and the stiffness-aware index (SAI) \cite{liang2021stiffness,huang1997adaptive} $\zeta_{\text{SAI}}$. Specifically, for the dynamics of the state $\mathbf{u}(t)$ in Eq.(\ref{eqn:ode}), the SI at the state $\mathbf{u}(t) \equiv u^t$ is defined by $\zeta_{\text{SI}}(u^t) = \max(|\mathbf{Re}(\lambda_i)|),$
% \begin{equation}
%     \zeta_{\text{SI}}(u^t) = \max(|\mathbf{Re}(\lambda_i)|),
%     \label{eq:si}
% \end{equation}
where $\lambda_i$ is the eigenvalue of the Jacobian matrix at $u^t$ for the right-hand side of Eq.(\ref{eqn:ode})~\cite{aiken1985stiff}. The maximum eigenvalue (real part) of the Jacobian matrix represents the speed of the change for the solution. %Let consider a simple linear system for Eq.(\ref{eqn:ode}), \textit{i.e.} the equation $\frac{d \mathbf{u}}{dt}=\mathbf{A}\mathbf{u}$, and $\mathbf{A}\in \mathbb{R}^{n\times n}$ is a diagonalizable matrix with eigenvalues $\{\lambda_i\}_{i=1}^n$, corresponding eigenvectors $\{\mathbf{v}_i\}_{i=1}^n$ and $\mathbf{Re}(\lambda_i)<0$ for $i=1,..., n$. This system have the explicit solution $\mathbf{u}(t)=\sum_{i=1}^nc_i\mathbf{v}_i e^{\lambda_it}$, 
In some data-driven settings, the analytic expression of the right-hand side of Eq.(\ref{eqn:ode}), \textit{i.e.,} $\mathbf{f}$, is unknown. In this case, the SAI is considered since the SAI does not require $\mathbf{f}$ to be known, and the SAI can be viewed as the proxy of the SI \cite{liang2021stiffness}. For the discrete observation data $\{u^{t_i}\}_{i-1}^N$ from a given system, the SAI at $u^{t_i}$ can be defined as
\vspace{-5pt}
\begin{equation}
        \zeta_{\text{SAI}}(u^{t_i}) = \frac{1}{\|u^{t_{i}}\|_2}\big\|\frac{u^{t_{i+1}}-u^{t_i}}{t_{i+1}-t_i}\big\|_2.
    \label{eq:sai}
\end{equation}
From Eq.(\ref{eq:sai}), the norm of the finite difference $\frac{u^{t_{i+1}}-u^{t_i}}{t_{i+1}-t_i}$ can approximate the instantaneous change of the unknown function $\mathbf{f}$ at $u^{t_i}$. The term $1/\|u^{t_{i}}\|_2$ is used to eliminate the bias of the magnitude of $\{u^{t_i}\}_{i-1}^N$ in different coordinate systems. If we only want to measure the relative stiffness information, such as the rank of the stiffness, we can use a simplified SAI, \textit{i.e.,} $\tilde{ \zeta}_{\text{SAI}}(u^{t_i}) = \frac{u^{t_{i+1}}-u^{t_i}}{t_{i+1}-t_i}$, as a new kind of stiffness information measurement for analysis~\cite{liang2021stiffness}. %When  $\zeta_{\text{SI}}$ or $\zeta_{\text{SAI}}$ is large, the solutions at some steps are stiff.
%引出没有归一化的形式

% To illustrate that SI reveals the fastest rate of change of state, we consider that Eq.(\ref{eqn:ode}) is a linear system with constant coefficients, namely, $\frac{\di \Bu}{dt}=\BA\Bu$, and $\BA\in \BR^{n\times n}$ is a diagonalizable matrix with eigenvalues $\{\lambda_i\}_{i=1}^n$ and corresponding eigenvectors $\{\Bv_i\}_{i=1}^n$. Then the solution of the linear system is $\Bu(t)=\sum_{i=1}^nc_i\Bv_i e^{\lambda_it}$. Let us suppose that $Re(\lambda_i)<0, i=1,..., n$. We have $e^{\lambda_it}\to \mathbf{0}$ as $t\to \infty$. Hence, $\max\{|Re(\lambda_i)|\}$ reveals the fastest speed of decaying to $\mathbf{0}$. If $\max\{|Re(\lambda_i)|\}$ is large, the integrator needs a small step size to reduce the local  truncation error.  

\begin{figure*}[t]
\centering
\includegraphics[width=1.0\linewidth]{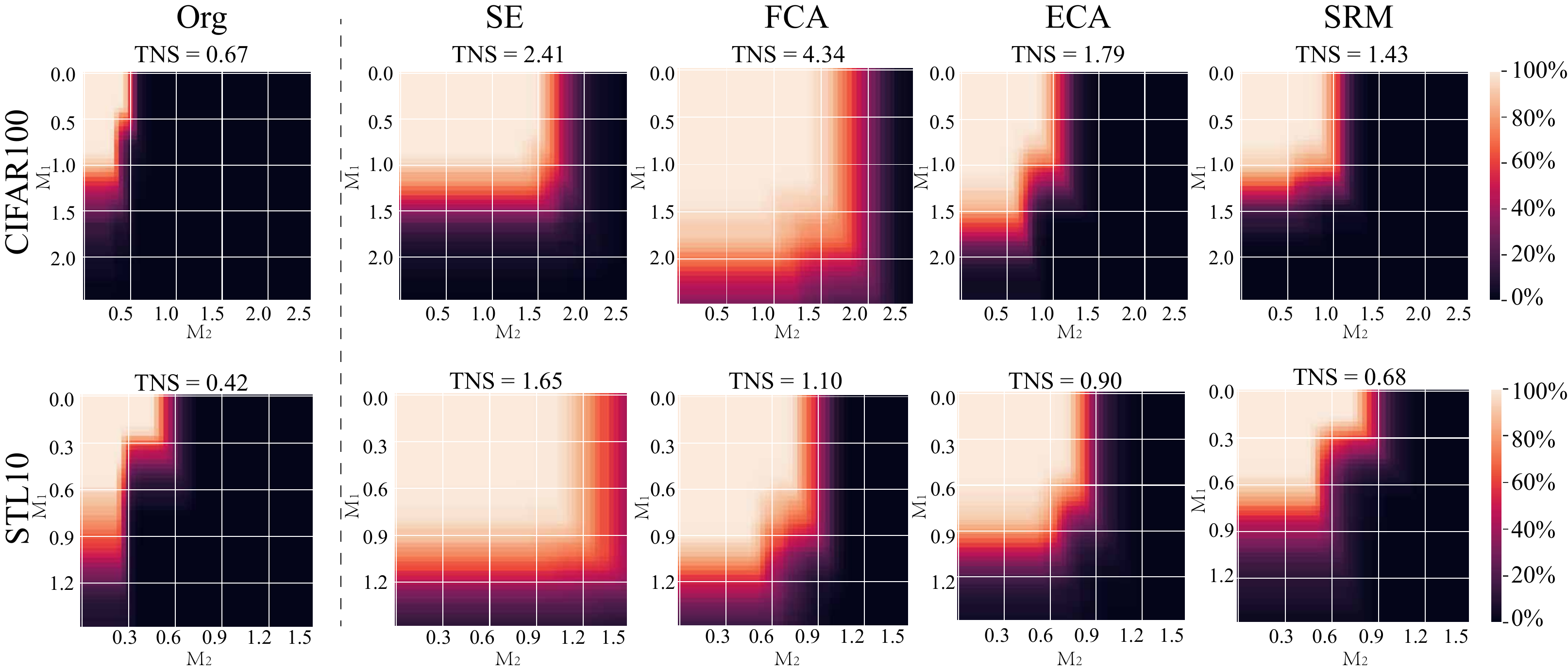}
\vspace{-15pt}
\caption{The quantitative visualization for SP by TNS for different models with residual blocks on CIFAR100 and STL10 datasets.} 

\label{fig:totalstiff}

\end{figure*}
\begin{figure*}[t]
\centering
\includegraphics[width=0.8\linewidth]{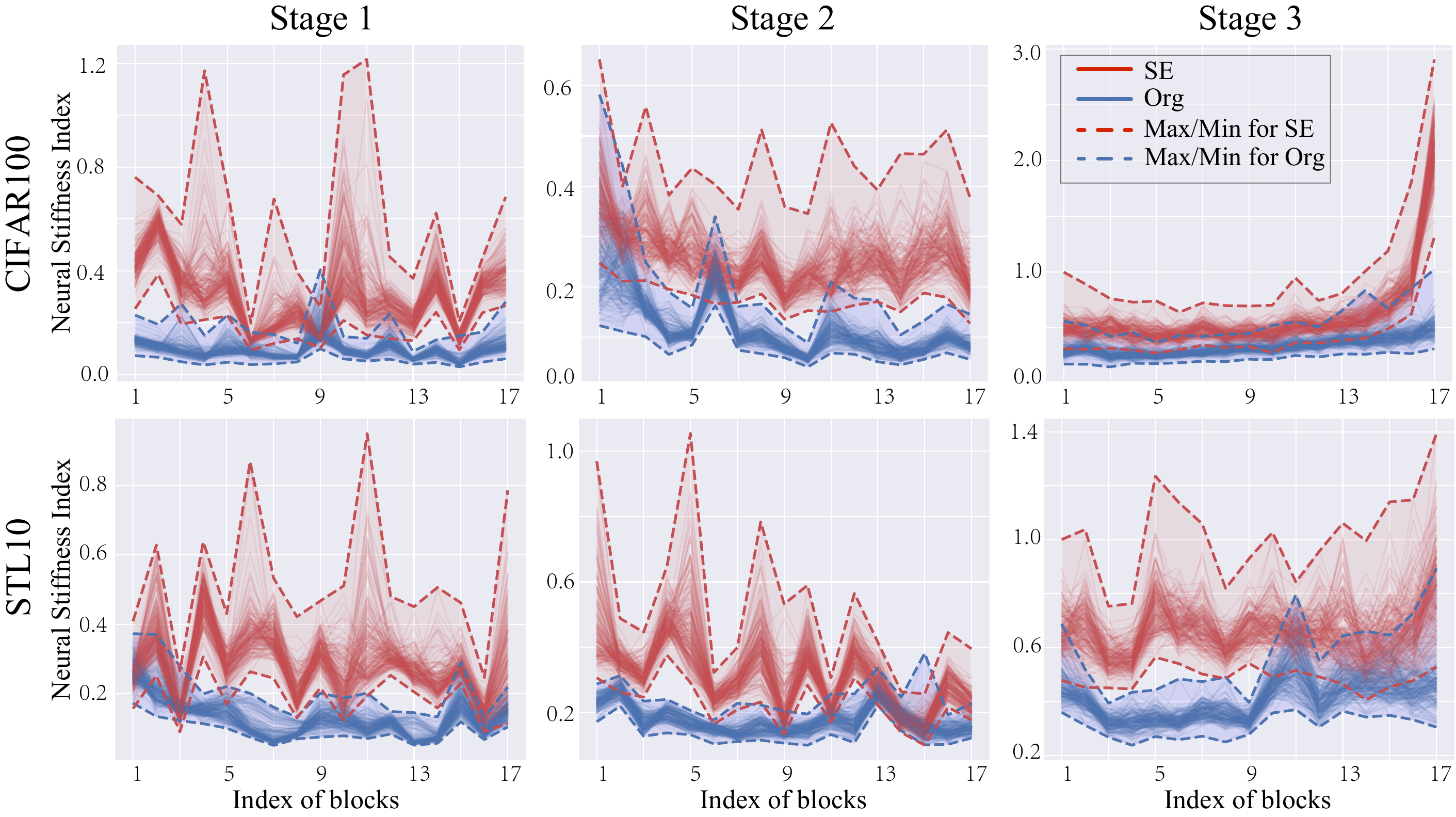}
% \vspace{-5pt}
\caption{The intuitive visualization for SP by NSI in the SENet164 and ResNet164 on CIFAR100 and STL10 datasets.} 
\vspace{-10pt}
\label{fig:stiffness_vis}
\end{figure*}

% \section{dfs}
\subsection{Measuring Stiffness Phenomenon~(SP) in NN}
\label{sec:stiff_neural}
In this section, based on the definition of stiffness in ODEs and the dynamical systems view of residual blocks presented in Section \ref{sec:preli}, we define the stiffness phenomenon~(SP) in the feature trajectory $[x_{1},x_{2},...]$ generated by Eq.(\ref{eq:resnet}) from two perspectives. Firstly, \ul{(1) from a local perspective}, we propose the Neural Stiffness Index (NSI) using the idea of Eq.(\ref{eq:sai}) and Definition \ref{define:stiffness_neural} to measure the SP based on the change of adjacent features. NSI can be used to visualize the SP qualitatively and intuitively. However, according to the related works about stiff ODEs \cite{huang1997adaptive,liang2021stiffness}, NSI needs to be greater than a certain threshold, \textit{i.e.}, the local change needs to be large enough, to be considered that the SP happens. However, the threshold varies in different scenarios, and it is hard to pick a unified threshold \cite{liang2021stiffness}. 

Therefore, secondly, \ul{(2) from a global perspective}, we propose a versatile metric called Total Neural Stiffness (TNS) using NSI to quantitatively measure the stiffness phenomenon of any neural networks on a given dataset. The TNS defined in Eq.(\ref{eq:total}) considers all threshold settings, including the relative threshold $\mu(1+M_1)$ and the absolute threshold $M_2$, $\mathbf{M} \in \mathbb{R}^+\times \mathbb{R}^+$. A larger TNS value indicates a more apparent stiffness phenomenon on the feature trajectory. The convergence of TNS is guaranteed by Theorem \ref{leamma:convge}.
Furthermore, from Eq.(\ref{eq:sai_neural}), the calculation of NSI requires the 2-norm of the feature. However, previous works \cite{li2016pruning,huang2020convolution,ding2019global,huang2021rethinking} show that the norm in the first block in each stage are usually extremely large and sensitive, which may affect the calculation of Eq.(\ref{eq:total}). Therefore, we will exclude these features while measuring TNS. % Further discussion about TNS is shown in Section \ref{sec:disc}.
%在本节中，根据在Section \ref{sec:preli}中关于stiffness在ODE的定义以及残差块的动力系统观点，我们可以从两个角度定义NN中 the feature trajectory $[x_{1},x_{2},...]$ generated by Eq.(\ref{eq:resnet}) 上的刚性现象: (1)从局部的角度，我们利用公式\ref{eq:sai}的思想，提出Neural Stiffness Index~(NSI)，从相邻特征的变化来描述刚性现象，如定义\ref{define:stiffness_neural}. NSI可以用于定性地且直觉地可视化刚性现象引起的特征快速变化现象。根据ODEs中刚性的定义\cite{huang1997adaptive}， NSI需要大于一定阈值才能认为局部的变化比较剧烈，从而可以认为是出现了刚性现象，但不同的场景下所需要的阈值是不一样的，很难给出一个统一的阈值\cite{liang2021stiffness}。因此进一步地，(2)从全局的角度，利用NSI，我们提出了a versatile metric, called Total Neural Stiffness (TNS), 可以用于定量地描述给定数据集下，任意神经网络的刚性现象。TNS在公式\ref{eq:total}中相当于为NSI考虑了所有阈值的情况，包括相对的阈值$\mu(1+M_1)$和绝对的阈值$M_2$这两种设定。如果TNS越大，那么说明该特征轨迹有很明显的刚性现象。 the convergence of TNS is guaranteed by Theorem \ref{leamma:convge}.  
 % As mentioned in Section \ref{sec:view}, a neural network with residual blocks can correspond to the numerical solution of ODE and thus in general we can also interpret the stiffness of the NN by SI. However, it is not easy to define the stiffness in NN by SI since the neural network is obtained in a data-driven manner and thus its corresponding equation is unknown. Instead, we can consider SAI to define the Neural Stiffness Index~(NSI) for the neural network as Definition \ref{define:stiffness_neural}, given the feature trajectory $[x_{1},x_{2},...]$ generated by Eq.(\ref{eq:resnet}). 
\begin{definition}\label{define:stiffness_neural}
(\underline{Neural Stiffness Index}) For the feature trajectories $x_{1},x_{2},x_{3}...,x_{L}$ generated by a neural network with $L$ residual blocks, \textit{i.e.,} $x_{t+1} = x_t + f(x_t;\theta_t)\cdot \Delta_t, t=0,1,..,L-1$. The Neural Stiffness Index~(NSI) at $x_t$ is 
\vspace{-5pt}
\begin{equation}
        \zeta_{\text{NSI}}(x_t) = \frac{1}{\|x_t\|_2}\big\|\frac{x_{t+1}-x_t}{\Delta_t}\big\|_2,
    \label{eq:sai_neural}
\end{equation}
e.g., for NN in Eq.(\ref{eq:resnet}), $\zeta_{\text{NSI}}(x_t) = \big\|x_{t+1}-x_t\big\|_2 / \|x_t\|_2$.

\end{definition}

%阈值 不通用 一版连接处不做考虑CWDA

% at t th block? 而不是xt?

\begin{definition}\label{define:stiffness_neural_plus}
(\underline{Total Neural Stiffness}) In Definition \ref{define:stiffness_neural}, the feature trajectory has the stiffness $\zeta_{\text{NSI}}(x_t;\mathbf{M})$ with degree $\mathbf{M} = (M_1,M_2) $ when $\exists t$ such that $\zeta_{\text{NSI}}(x_t) \geq \max(\mu(1+M_1),M_2),$
% \begin{equation}
%     \zeta_{\text{NSI}}(x_t) \geq \max(\mu(1+M_1),M_2),
%     \label{eq:stiff_total1}
% \end{equation}
where $\mu$ is the mean of all features' NSI from stage $S$ of $x_t$. For input $x_0$ sampled from test distribution $P(x_0)$, the Total Neural Stiffness is $ \iint_{\mathbf{M}} \delta(\mathbf{M}) d\mathbf{M}$, where
\vspace{-5pt}
\begin{equation}
     \delta(\mathbf{M}) = \mathbb{E}_{x_0\sim P(x_0)} \mathbf{I}_{\exists t, s.t. \zeta_{\text{NSI}}(x_t) \geq \max(\mu(1+M_1),M_2)},
    \label{eq:total}
\end{equation}
and $\mathbf{I}$ is the characteristic function, \textit{i.e.,} when $\zeta_{\text{NSI}}(x_t)$ not less than $\max(\mu(1+M_1),M_2)$, $\mathbf{I} = 1$, otherwise, $\mathbf{I} = 0$, $\mathbf{M} \in \mathbb{R}^+\times \mathbb{R}^+$.

\end{definition}

 %%%%%%%%%%%%%%%%%%%%%%%

% NSI provides us a method to measure whether a feature map $x_t$ from $t^{\rm th}$ block in a trajectory $[x_1,x_2,...]$ is stiff or not. If the NSI $\zeta_{\text{NSI}}(x_t)$ is large enough, the feature $x_t$ is considered stiff, otherwise, it is non-stiff. 
% Moreover, in Definition \ref{define:stiffness_neural_plus}, we propose a versatile metric, called Total Neural Stiffness (TNS), to measure whether a whole feature trajectory suffers from stiffness issue, and the convergence of TNS is guaranteed by Theorem \ref{leamma:convge}.
\begin{theorem}\label{leamma:convge}
 For $\delta(\mathbf{M})$ defined as Eq.(\ref{eq:total}) , the TNS $\iint_{\mathbf{M}} \delta(\mathbf{M}) d\mathbf{M}$ is convergent. See proof in the appendix.
\end{theorem}

% \begin{proof}
% 	(See Appendix A).\qedhere
% \end{proof}

% The reason for defining the TNS as Definition \ref{define:stiffness_neural_plus} are two-fold: (1) if the feature $x_t$ from $t^{\rm th}$ block is stiff, then there should be a large difference between its NSI $\zeta_{\text{NSI}}(x_t)$ and the average NSI $\mu$ from the whole stage. Therefore, $\zeta_{\text{NSI}}(x_t)$ should be large than a threshold $\mu(1+M_1)$, $M_1 \in \mathbb{R}_+$; (2) The threshold $\mu(1+M_1)$ is only a related threshold, which is not enough for measuring stiffness. Thus we will need to set an absolute threshold like $M_2 \in \mathbb{R}_+$. For example, let's consider standard normal distribution $N(0,1)$. Due to the three-sigma rule \cite{pukelsheim1994three}, although most of the sampled values in this distribution are concentrated around the mean value $\mathbf{0}$, it is always possible to sample some values that are extremely far from the mean value. However, the change of the sampled values may be smooth and these values may not have rapid variation. Therefore, we can follow the previous work \cite{liang2021stiffness} and set an absolute threshold $M_2 \in \mathbb{R}_+$ as shown in Definition \ref{define:stiffness_neural_plus}. After that, we use $ \iint_{\mathbf{M}} \delta(\mathbf{M}) d\mathbf{M}$ to measure the stiffness for a given neural network with residual blocks. If TNS is large enough, we can consider the given trajectory has a stiffness issue. 

%不同的stage有不同的维度

\subsection{The Ground Truth (GT) Trajectory}
\label{sec:gtt}
%无穷多条

To explore the properties of the SP in the NN, it is desirable to have a ``ground truth" such that we can analyze the relationship between the properties of the SP and the model's performance to propose specific improvements for the model. 
Given a dataset $\mathcal{D}(x,y)$ and a network $\mathcal{A}(\theta_0,s)$ with residual blocks, where $\theta_0$ is initialization parameters and $s$ denotes training setting~(e.g., learning rate, weight decay, etc.). 
We consider some feature trajectories whose corresponding model $\mathcal{A}(\theta_0,s)$ have supremum performance as GT trajectories in Definition \ref{define:gt}.
%有界metricS

\begin{definition}\label{define:gt}
(\underline{GT trajectory}) For the dataset $\mathcal{D}(x,y)$, a task-oriented performance metrics $\kappa$, and a residual neural network $\mathcal{A}(\theta_0,s)$ with initialization parameters $\theta_0$ and training setting $s$. After training, there are an infinite number of feature trajectories $[x_{i1},x_{i2},...],i=1,2,...,\infty$, from a given input $x_0$ to the corresponding output $y_0$. The GT trajectories are the feature trajectories whose performance of $\mathcal{A}(\theta_0,s)$ on $\mathcal{D}(x,y)$ can reach
\begin{equation}
    \sup_{\theta_0,s} \kappa(\mathcal{A}(\theta_0,s),\mathcal{D}(x,y)).
    \label{eq:sup}
\end{equation}
\end{definition}
%In fact, the GT trajectories exists according to Definition \ref{define:gt}.
First, the GT trajectories introduced in Definition \ref{define:gt} exist. Let's consider the non-empty set $K$ whose elements are the metrics $\kappa$ under all initialization parameters $\theta_0$ and training setting $s$. Note that the task-oriented performance metrics $\kappa$ are usually bounded in various deep learning tasks, especially in supervised learning, \textit{e.g.}, the upper bound for the metrics of the classification task is 100\%, so $\kappa(\mathcal{A}(\theta_0,s),\mathcal{D}(x,y))<+\infty$. According to the well-known theorem~\cite{courant1965introduction} that every non-empty subset $\subset \mathbb{R}$ which has an upper bound has the supremum, thus the set $K$ has the supremum, which means the GT trajectories exist. 

However, this kind of trajectory given an input $x_0$ may not be unique, which will be shown in Appendix. Moreover, as mentioned in Section \ref{sec:stiff_neural}, the feature trajectory is data-driven, and the corresponding analytic expression of the right-hand side of Eq.(\ref{eqn:ode}) in a NN is unknown. Thus it is infeasible to obtain the analytical form of the GT trajectories, and instead, trajectories from the model with high enough performance, like some advanced self-attention networks, can be seen as the proxies to empirically approach the properties of GT trajectories. Therefore, we take ResNet164 as the backbone and select four high-performance self-attention networks, \textit{i.e.}, SENet \cite{hu2018squeeze}, FCANet \cite{qin2021fcanet}, ECANet \cite{wang2020eca}, and SRMNet \cite{lee2019srm}, to approximately analyze the properties of GT trajectories. In Fig.\ref{fig:totalstiff}, we show the TNS for each network with different $M_1$, $M_2$. For the CIFAR100 and STL10 datasets, from the results of all high-performance networks in Fig.\ref{fig:totalstiff}, we observe that the GT trajectories have a significantly large TNS. In other words, for most inputs, the GT trajectories have the SP. Although the TNS of the original residual neural network is relatively small, it can still measure the SP to some extent.
 % For the results of ``Org", it shows that the original residual neural network can also measure TNS to some extent.
Moreover, as we can empirically observe that SP widely exists in these high-performance networks, we conjecture that the existence of SP is an intrinsic property of GT trajectories (More discussions are shown in Section \ref{sec:disc}). Thus if a NN structure can better estimate the stiffness information to measure and capture the SP, \textit{e.g.,}, with high TNS, then such a structure can generate feature trajectories that are closer to the GT trajectories and thus achieve high performance. Otherwise, if a NN structure can not inherently measure the stiffness information, it may produce large deviations between its feature trajectory and the GT trajectories. Such a deviation will gradually accumulate with the forward process of NN, leading to poor prediction. Lastly, in Fig.\ref{fig:stiffness_vis}, we provide the intuitive visualization of the SP by NSI in Definition \ref{define:stiffness_neural}. Obviously, on both CIFAR100 and STL10 datasets, we can observe that the GT trajectory measured by the SENet has significant and rapid oscillations in each stage, leading to the SP. The trajectories measured by ``Org" are relatively smooth but also have some oscillations that are consistent with the observation in Fig.\ref{fig:totalstiff}.

 %我们选取了四种注意力网络 SENet, DIANet, ECANet and SRMNet来分析GT trajectories。在图2中，我们给出了不同网络下的B
 
% Therefore, in order to explore the stiffness of this kind of trajectory, we use advanced self-attention networks with high performance, including SENet, DIANet, ECANet and SRMNet as proxies to approximate the stiffness of GT trajectories. As shown in Fig.\ref{fig:totalstiff}

%  Note that the Sp in transition layer are relatively small, and the transition layer refers to the layer where the dimensions of the filter change, like the layer between stage 1 and stage 2 of a ResNet. The reason for this phenomenon may be that the layers in these areas are sensitive. It is interesting but will not greatly impact the structural similarity of the whole pruned network. The similar observations are shown in Fig.~2 in \cite{ding2019global}, Fig.~6 and Fig.~10 in \cite{li2016pruning}.

%cwda 去掉头

% stifness是本身的性质

\subsection{Self-attention Mechanism~(SAM) and Stiffness-aware Step Size Adaptor}
\label{sec:self_and_step}

In this section, we introduce the self-attention mechanism~(SAM) and reveal the role it plays in improving model performance based on the dynamical perspective of the neural networks introduced in the above sections. For the analysis in the main paper, we use the channel attention neural networks~\cite{hu2018squeeze,wang2020eca,huang2020dianet} as an example, and the transformer-based self-attention models~\cite{liu2021swin,vaswani2017attention} will be discussed in Appendix. For the channel self-attention networks, they can be written as follows by comparing with Eq.(\ref{eq:resnet}):
\vspace{-5pt}
\begin{equation}
    \hat{x}_{t+1} = x_t + \underbrace{f(x_t;\theta_t)}_{\text{Feature map}} \otimes \underbrace{\mathbf{F}(f(x_t;\theta_t);\phi_t)}_{\text{Attention value}},
\label{eq:att}
\end{equation}
where $\otimes$ is the Hadamard product and $\mathbf{F}(\cdot;\phi_t)$ is the self-attention module with learnable parameters $\phi_t$ in $t^{\rm th}$ blocks based on different attention methods. For example, in SENet, $\mathbf{F}(\cdot;\phi_t) = \mathbf{W_{t1}}(\text{ReLU}(\mathbf{W_{t2}(\cdot)}))$, where $\mathbf{W_{t1}}\in \mathbb{R}^{d\times r}$ and $\mathbf{W_{t2}}\in \mathbb{R}^{r\times d}$ are learnable matrices, $r<d$.

\begin{figure*}[t]
\centering
\includegraphics[width=0.9\linewidth]{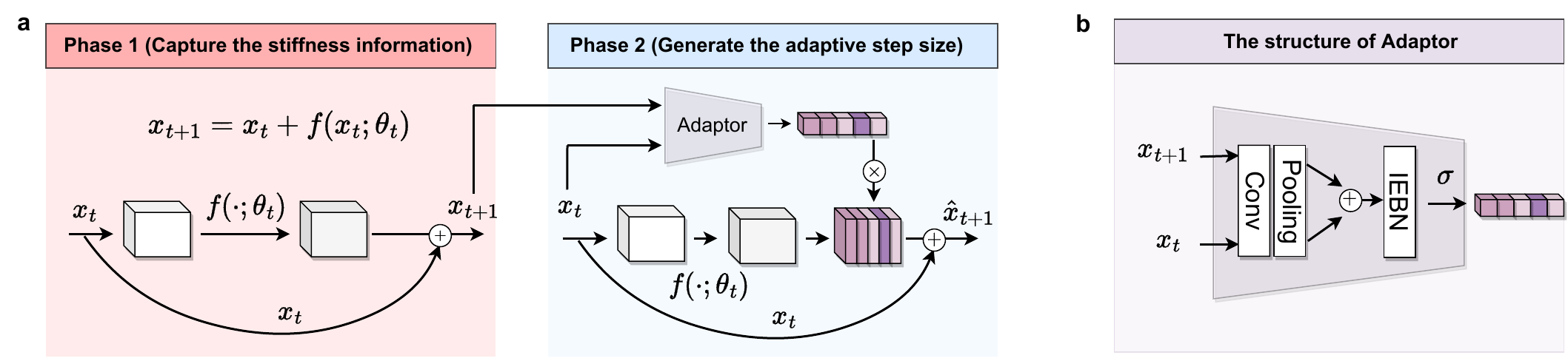}
        \vspace{-0.1cm}
\caption{The architecture of StepNet. \textbf{a,} the forward process of StepNet includes two phases. \textbf{b,} the structure of Adaptor. ``Conv" is group convolution, ``Pooling" is global average pooling and ``IEBN" \cite{liang2020instance} is the combination of batch normalization and a linear transformation ``IE" from \cite{huang2022layer}. $\sigma$ is \text{Sigmoid} activation function. More training details are provided in Appendix.} 
% \vspace{-0.2cm}
\label{fig:arch}
% \vspace{-0.3cm}
\end{figure*}
%都是小于1 不同的情况有不同的step的情况，我们以xxx为例
\begin{table*}[htbp]
\small
  \centering
  \vspace{-5pt}
  \resizebox*{0.9\linewidth}{!}{
        \begin{tabular}{lccccccccc}
    \toprule
    \multirow{2}[4]{*}{Model} & \multicolumn{3}{c}{CIFAR10} & \multicolumn{3}{c}{CIFAR100} & \multicolumn{3}{c}{STL10} \\
\cmidrule{2-10}          & \#P(M)  & top-1 acc. (\%) &  acc. $\uparrow$ & \#P(M) & top-1 acc. (\%) &  acc. $\uparrow$ & \#P(M)  & top-1 acc. (\%) &  acc. $\uparrow$\\
    \midrule
    ResNet~\cite{he2016deep} & 1.70   & 93.35 $\pm$ 0.18 & - &1.73 & 74.30 $\pm$ 0.30 &-   & 1.70   & 82.66 $\pm$ 1.05 & -\\
    +SE~\cite{hu2018squeeze} & 1.91    & 94.26 $\pm$ 0.14 &0.91 &  1.93    & 75.25 $\pm$ 0.31 &0.95 & 1.91   & 83.96 $\pm$ 0.92 &1.30 \\
    +CBAM~\cite{woo2018cbam} & 1.92   & 93.91 $\pm$ 0.14 &0.56 & 1.94    & 74.68 $\pm$ 0.22 & 0.38   & 1.92    & 84.11 $\pm$ 0.49 &1.45 \\
    +ECA~\cite{wang2020eca} & 1.70    & 94.15 $\pm$ 0.23 &0.80 & 1.73    & 74.41 $\pm$ 0.27 &0.11 & 1.70   & 83.89 $\pm$ 1.21 &1.23 \\
    % ResNet164-SpaNet~\cite{guo2020spanet} & 3.83  & 94.31 & 3.86  & 75.70 & 3.83  &  \\
    +SEM~\cite{zhong2022switchable} & 1.95    & 94.52 $\pm$ 0.11  &1.17 &  1.97   & 76.49 $\pm$ 0.26 &2.19 & 1.95    & \underline{85.73 $\pm$ 0.16} &\underline{3.07}\\
    +SRM~\cite{lee2019srm} & 1.74    & 94.59 $\pm$ 0.12 & 1.24 &  1.76    & 75.49 $\pm$ 0.31 &1.19 & 1.74    & 84.73 $\pm$ 0.12 &2.07 \\
    +FCA~\cite{qin2021fcanet} & 1.91    & 94.66 $\pm$ 0.12 &1.31 & 1.93    & \underline{76.52 $\pm$ 0.19} &\underline{2.21} & 1.91    & 85.34 $\pm$ 0.25  &2.68\\
    +IEBN~\cite{liang2020instance} & 1.73    & \underline{94.70 $\pm$ 0.13} &\underline{1.35} &  1.75    & 76.32 $\pm$ 0.14 &2.02 & 1.73     & 85.13 $\pm$ 0.04 &2.47 \\
    \rowcolor{Gray}+StepNet (Ours) & 1.75     & \pmb{95.14 $\pm$ 0.24}&\pmb{1.79}  & 1.77     & \pmb{77.04 $\pm$ 0.22} &\pmb{2.74} & 1.75     & \pmb{86.08 $\pm$ 0.11} &\pmb{3.42} \\
    \bottomrule
    \end{tabular}%
  }
  \caption{The classification accuracy on CIFAR10, CIFAR100, and STL10. ``\#P(M)" means the number of parameters (million). Bold and underline indicate the best results and the second best results, respectively. }
  \vspace{-10pt}
  \label{tab:main}%
\end{table*}%

Compared with the forward numerical method mentioned in Section \ref{sec:stiff_math}, the attention value $\mathbf{F}(f(x_t;\theta_t);\phi_t)$ in Eq.(\ref{eq:att}) can be referred to as the step size in solving ODEs, and the original residual neural networks can be rewritten as
\vspace{-5pt}
\begin{equation}
    \label{org step size}
    x_{t+1} = x_t + f(x_t;\theta_t)\otimes \Delta t,
\end{equation}
where the step size $\Delta t = \mathbf{1}$. By comparing Eq.(\ref{eq:att}) and Eq.(\ref{org step size}), we can readily find that the attention value generated by the SAM exactly serves as an adaptive step size, \textit{i.e.}, $\Delta t = \mathbf{F}(f(x_t;\theta_t);\phi_t)$. Moreover, since the last layers of various self-attention modules are usually $\text{Sigmoid}$ function or $\text{Softmax}$ function, thus the attention values in Eq.(\ref{eq:att}) are less than 1, \textit{i.e.}, $\Delta t = \mathbf{F}(f(x_t;\theta_t);\phi_t) < 1$. This implies that with SAM, we can provide a smaller and more flexible step size than that from the original residual neural network. Now we show that the step size generated by the SAM is also stiffness-aware.
As mentioned in Section \ref{sec:stiff_math}, we can use $ (x_{t+1} - x_t)/\Delta t$
to measure the stiffness information in feature trajectory at $x_t$. Then for the the original residual neural networks in Eq.(\ref{eq:resnet}), we have
\vspace{-5pt}
\begin{equation}
     f(x_t;\theta_t) = \frac{x_{t+1}-x_t}{\Delta t}|_{\Delta t = 1}\equiv \check{ \zeta}_{\text{NSI}}(x_t),
    \label{eq:stiffbyorg}
\end{equation}
and hence the $f(x_t;\theta_t)$ can be regarded as a kind of coarse stiffness information with step size $\Delta t =1 $, which is consistent with the discussions of Fig.\ref{fig:totalstiff} and Fig.\ref{fig:stiffness_vis} in Section \ref{sec:gtt}. For the SAM, we have
\vspace{-5pt}
\begin{equation}
\begin{aligned}
\mathbf{F}(f(x_t;\theta_t);\phi_t) &= \mathbf{F}(\underbrace{\frac{1}{\Delta t}(x_{t+1} - x_t)|_{\Delta t = 1}  }_{\text{Coarse stiffness information}};\phi_t)\\
&=\mathbf{F}(\check{ \zeta}_{\text{NSI}}(x_t);\phi_t).\\
\end{aligned}
\label{eq:adapt}
\end{equation}	
From Eq.(\ref{eq:adapt}), we can summarize how the self-attention module helps the performance of the original residual neural networks: \ul{(1) Capture the stiffness information.} the self-attention module $\mathbf{F}(\cdot;\phi_t)$ take the accessible and coarse stiffness information $f(x_t;\theta_t)$ from Eq.(\ref{eq:stiffbyorg}) as input. Then as shown in Fig.\ref{fig:totalstiff} and Fig.\ref{fig:stiffness_vis}, the self-attention module can refine this coarse information to obtain a finer estimation of stiffness information; \ul{(2) Generate the adaptive step size}. Based on this finer estimation, the module $\mathbf{F}(\cdot;\phi_t)$ outputs suitable attention values $\mathbf{F}(f(x_t;\theta_t);\phi_t)$ to adaptively measure the SP in the neural network, which means the SAM can enhance the representational ability of NN. For example, if the feature trajectory at $x_t$ needs to measure a large NSI, from Eq.(\ref{eq:sai_neural}), the attention value $\mathbf{F}(f(x_t;\theta_t);\phi_t)$ can be small to get the large
$\zeta_{\text{NSI}}(x_t) = \frac{1}{\|x_t\|_2}\big\|\frac{\hat{x}_{t+1}-x_t}{\mathbf{F}(f(x_t;\theta_t);\phi_t)}\big\|_2$.

\subsection{Theoretic-inspired Approach: StepNet}
\label{sec:method}

From Section \ref{sec:self_and_step} and Eq.(\ref{eq:adapt}), we know that the ability to properly estimate the stiffness information is essential for the performance of the self-attention module. Thus if we want to obtain better model performance, we can consider estimating other accessible and better stiffness information in the self-attention module. Now we introduce a better self-attention formulation to capture better stiffness information, which is motivated by the asymptotic analysis between SI and SAI as follows. 
In Section \ref{sec:stiff_math}, SAI is used as a proxy for the versatile index of stiffness~(SI) to measure stiffness information in ODEs for tackling the computational difficulties of SI in data-driven problems. In Theorem \ref{theo:saivssi}, we first show how SAI can approximate the SI.

\begin{theorem}\label{theo:saivssi}
 For an ODE $\text{d}\mathbf{u}(t)/\text{d}t = \mathbf{f}[\mathbf{u}(t)]$ defined at Eq.(\ref{eqn:ode}), if the Jacobian matrix $\mathbf{J}_{u^t}$ at $u^t$ is a $n\times n$ symmetric real matrix and $\{\lambda_i\}_{i=1}^n$ are its $n$ distinct eigenvalues, and $\mathbf{Re}(\lambda_i)<0,i=1,2,...,n$, then
 \begin{equation}
     \zeta_{\text{SAI}}(u^{t})  \approx   \zeta_{\text{SI}}(u^t) \cdot \sqrt{c + Q[\zeta_{\text{SI}}(u^t)]},
     \label{eq:approx_si_sai}
 \end{equation}
 where $c$ is a constant and $Q(\cdot)$ is a function with respect to $\zeta_{\text{SI}}(u^t)$ and when $\zeta_{\text{SI}}(u^t)$ is large enough, $Q[\zeta_{\text{SI}}(u^t)]$ converges to a 0. See proof in the appendix.
\end{theorem}

% \begin{proof}
% 	(See Appendix).\qedhere
% \end{proof}

From Eq.(\ref{eq:approx_si_sai}) we can see that $\sqrt{c + Q[\zeta_{\text{SI}}(u^t)]}$ tends to $\sqrt{c }$ when SI is large and then $\zeta_{\text{SAI}}(u^{t}) \propto \zeta_{\text{SI}}(u^t)$, \textit{i.e.}, $\zeta_{\text{SAI}}(u^{t})$ and $\zeta_{\text{SI}}(u^t)$ are positively correlated, which make the Eq.(\ref{eq:stiffbyorg}) constructed from SAI provide relatively accurate stiffness information. However, if the SI is not large enough, the SAI and SI will be nonlinearly related, in other words, the stiffness information provided by Eq.(\ref{eq:stiffbyorg}) may not be precise enough and need further refinement.

Motivated by the above analysis, we propose a novel self-attention network called StepNet.
As discussed in Section \ref{sec:stiff_math}, the stiffness in the ODE may measure a changing trend like the rapid variation in solution, which implies that the stiffness information should be measured by two adjacent states. Therefore, for Eq.(\ref{eq:adapt}), we use a data-driven function $\tilde{\mathbf{F}}(x_{t+1}, x_t;\tilde{\phi}_t) = \tilde{\mathbf{F}}(x_t + f(x_t;\theta_t), x_t;\tilde{\phi}_t)$ to replace $\mathbf{F}(x_{t+1}- x_t;\phi_t)$, such that the adaptive step size generated in our StepNet is based on the intrinsic relation between two adjacent states $x_{t+1}$ and $x_{t}$ to better model the stiffness information. Thus, the Eq.(\ref{eq:att}) can be rewritten as 
\vspace{-5pt}
\begin{equation}
\begin{aligned}
\hat{x}_{t+1} &= x_t + f(x_t;\theta_t) \otimes \mathbf{F}(f(x_t;\theta_t);\phi_t)\\
&\gets x_t + f(x_t;\theta_t) \otimes \tilde{\mathbf{F}}(x_t + f(x_t;\theta_t), x_t;\tilde{\phi}_t).\\
\end{aligned}
\label{eq:ours}
\end{equation}	
From Eq.(\ref{eq:ours}), the calculation of our StepNet has two phases: (1) In Fig.\ref{fig:arch} (a), we first estimate a coarse $(t+1) ^{\rm th}$ feature map $x_{t+1}= x_t + f(x_t;\theta_t)$ by Eq.(\ref{eq:resnet}); (2) After that, the Adaptor $\tilde{\mathbf{F}}( \cdot,\cdot;\tilde{\phi}_t)$ take both the $x_t$ and $x_{t+1}$ as input to generate the adaptive attention values, which can better measure the stiffness information to generate finer step size for better capturing the SP to enhance the representational ability of the model and boost the performance. The network architecture of the Adaptor is shown in Fig.\ref{fig:arch} (b) and more training details are provided in Appendix.

\begin{figure*}[t]
\centering
\includegraphics[width=0.95\linewidth]{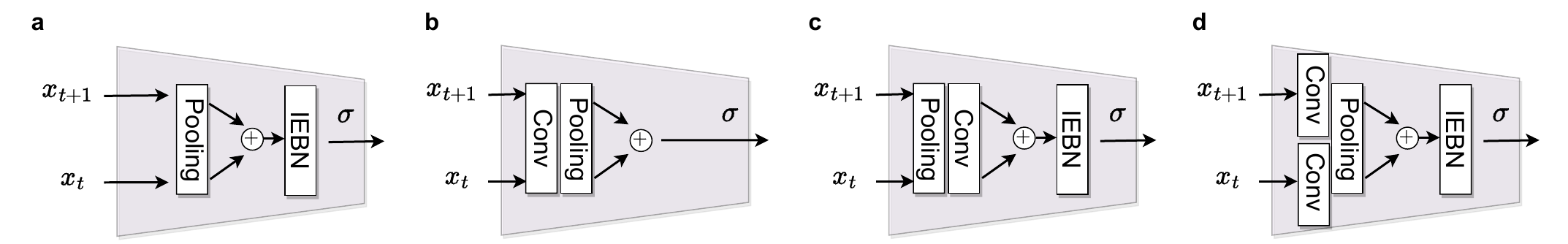}
        \vspace{-0.1cm}
\caption{Different structures of Adaptor.} 
\vspace{-5pt}
\label{fig:ablation}
\end{figure*}

\section{Experiment}

In this section, we use several popular vision benchmarks to verify the effectiveness of the proposed StepNet, including image classification and object detection. All experiments are verified 5 times with random seeds, and the average performances with standard deviations are reported. The experimental settings can be found in Appendix.

% \subsection{Image classification}
% \label{sec:cls}
\textbf{Image classification}. We compare the proposed StepNet with several existing self-attention modules on four datasets for image classification. These four datasets are CIFAR10~\cite{krizhevsky2009learning}, CIFAR100~\cite{krizhevsky2009learning}, STL-10~\cite{pmlr-v15-coates11a} and ImageNet~\cite{NIPS2016_90e13578}, and the results of these datasets are listed in Table \ref{tab:main} and \ref{tab:imagenet}, which show that the StepNet improves the accuracy significantly over the original networks and consistently compared with other existing self-attention modules under different datasets and backbones.

\textbf{Object detection}. We further conduct experiments for object detection tasks on the MS COCO dataset with Faster R-CNN, and the results are shown in Table \ref{tab:object}, where our StepNet outperforms the baseline models by 2.9\% and 3.0\% in terms of AP on ResNet50 and ResNet101, respectively. Moreover, the StepNet also achieves good enough performance improvements in other object detection metrics. All the results show that our new understanding of the SAM can effectively help us propose a well design for SAM.

% \tiny
\begin{table}[t]
\small
  \centering
    \begin{tabular}{lrr}
    \toprule
     \multicolumn{1}{c}{\textbf{Model}} & \multicolumn{1}{c}{\textbf{ResNet34-acc.}} & \multicolumn{1}{c}{\textbf{ResNet50-acc.}}\\
    \midrule
     Org &  73.99$\pm$ 0.04 & 76.02$\pm$ 0.09\\
     SE~\cite{hu2018squeeze}    & 74.37$\pm$ 0.10 & 76.62$\pm$ 0.02\\
     ECA~\cite{wang2020eca}    & 74.56$\pm$ 0.08 & 77.07$\pm$ 0.09\\
     CBAM~\cite{woo2018cbam}  & 74.89$\pm$ 0.11 & 76.38$\pm$ 0.10\\
     SRM~\cite{lee2019srm}   &  74.79$\pm$ 0.04 & 76.49$\pm$ 0.03\\
     FCA~\cite{qin2021fcanet}   & \pmb{75.08$\pm$ 0.19} &  \underline{77.22$\pm$ 0.15}\\
     SGE~\cite{li2019spatial} &  74.50$\pm$ 0.03 & 77.13$\pm$ 0.06\\
    \rowcolor{Gray} StepNet (Ours) & \underline{75.01$\pm$ 0.09} &  \pmb{77.52$\pm$ 0.04}\\
    % \midrule
    % ResNet50 & Org & 76.02$\pm$ 0.09 \\
    % ResNet50 & SE~\cite{hu2018squeeze}    & 76.62$\pm$ 0.02 \\
    % ResNet50 & ECA~\cite{wang2020eca}    & 77.07$\pm$ 0.09 \\
    % ResNet50 & CBAM~\cite{woo2018cbam}  & 76.38$\pm$ 0.10 \\
    % ResNet50 & SRM~\cite{lee2019srm}   & 76.49$\pm$ 0.03 \\
    % ResNet50 & SPA~\cite{guo2020spanet}   &  77.02$\pm$ 0.12\\
    % ResNet50 & DIA~\cite{huang2020dianet} & \pmb{77.52$\pm$ 0.06} \\
    % \rowcolor{Gray}ResNet50 & StepNet (Ours) &  \underline{\textit{77.12$\pm$ 0.04}}  \\

    \bottomrule
    \end{tabular}%
    \caption{The classification performance on ImageNet. The best and the second best results of each setting are marked in bold and \underline{\textit{italic}} fonts, respectively.}
  \label{tab:imagenet}%
\end{table}%

\begin{table}[t]
\small
  \centering
 \resizebox{0.95\columnwidth}{!}{
    \begin{tabular}{lrrrrrr}
    \toprule
    \textbf{Model} & \multicolumn{1}{l}{\textbf{AP }} & \multicolumn{1}{l}{\textbf{AP$_{50}$}} & \multicolumn{1}{l}{\textbf{AP$_{75}$}} & \multicolumn{1}{l}{\textbf{AP$_{S}$}} & \multicolumn{1}{l}{\textbf{AP$_{M}$}} & \multicolumn{1}{l}{\textbf{AP$_{L}$}} \\
    \midrule
    \multicolumn{1}{l}{ResNet50} &   36.9   &    57.5   &   39.4    &    22.1   &   39.6  &    46.4  \\
     +SE~\cite{hu2018squeeze}    &  38.3     &  60.1     & 41.5      &  23.0     &  42.1     &    \underline{49.2}  \\
    +ECA~\cite{wang2020eca}    &  38.5     &  \underline{60.4}     &  41.5     &  \underline{23.2}     &  \underline{42.2}     &  48.9  \\
    +SGE~\cite{li2019spatial}   &  38.9     &  60.0     &  \pmb{42.0}     &  \underline{23.2}     &  41.9     &  49.0  \\
    \rowcolor{Gray}+StepNet (Ours)     &  \pmb{39.8}     &  \pmb{61.2}     & \underline{41.9}     &  \pmb{23.8}     & \pmb{43.1}      & \pmb{51.0} \\
    \midrule
    \multicolumn{1}{l}{ResNet101} & 38.9      & 60.3      & 41.4      & 22.1      & 43.9      & 50.0  \\
    +SE~\cite{hu2018squeeze}    &  39.8     &    61.2   &    43.4   &    23.6   &    44.2   &    51.1  \\
    +ECA~\cite{wang2020eca}    &  \underline{40.4}     &    \pmb{62.7}   &    \pmb{44.5}   &    \underline{24.1}   &    \underline{45.1}   &    \underline{51.8}  \\
    +SGE~\cite{li2019spatial}   &  40.2     &    61.4   &    43.2   &    \underline{24.1}   &    44.5   &    51.2  \\
    \rowcolor{Gray}+StepNet (Ours) & \pmb{41.9}      & \underline{62.6}      &  \underline{44.4}     &  \pmb{25.4}     &  \pmb{46.8}     &  \pmb{53.2} \\
    \bottomrule
    \end{tabular}%
    }
    \caption{The object detection performance on MS COCO. The best and the second best results of each setting are marked in bold and \underline{\textit{italic}} fonts, respectively. 
 %  “AP”, “AP$_{S}$”,
 % “AP$_{M}$”, and “AP$_{L}$”: averaged AP for overall, small, medium, and large scale objects, respectively, at [50\%, 95\%] IoU interval with step as 5\%, “AP$_{50}$” and “AP$_{75}$”: AP at IoU as 50\% and 75\%, respectively. {\color{red}{Red}} color means the accuracy can be improved with different modules.
 }
    \vspace{-10pt}
  \label{tab:object}%
\end{table}%


\begin{figure*}[t]
\centering
\includegraphics[width=1.0\linewidth]{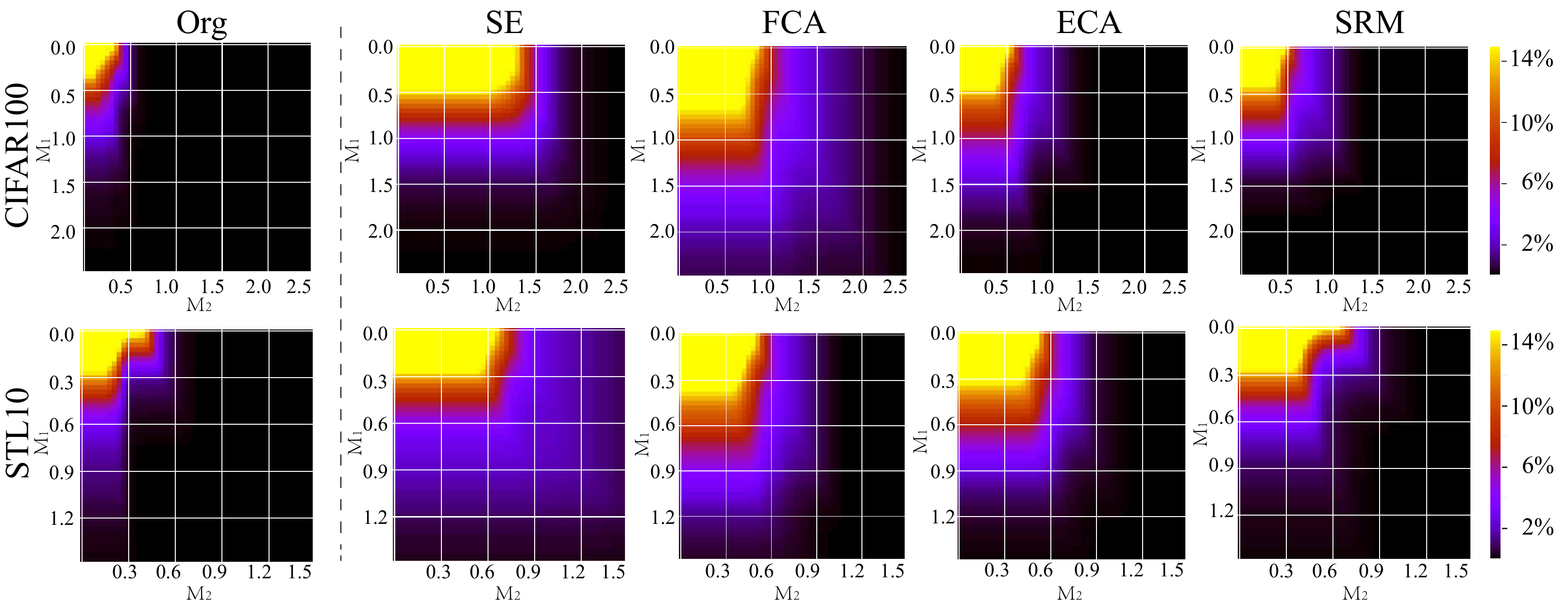}
\vspace{-15pt}
\caption{The stiffness proportion for different models with residual blocks on CIFAR100 and STL10.} 
\vspace{-5pt}
\label{fig:totalratio}

\end{figure*}

\subsection{Ablation Study and Discussions}
\label{sec:disc}
Now we further explore the property of the StepNet and the stiffness phenomenon in neural networks.

\textbf{The structure of StepNet.} In Section \ref{sec:self_and_step} we summarize that the self-attention mechanism can help the model performance in two ways, \textit{i.e.}, the extraction of stiffness information and the generation of adaptive step sizes. In fact, these two ways correspond to (1) the input of the self-attention module and (2) the design of the module. For (1), in Table \ref{tab:ablation}, we explore the performance when $x_t$ or $x_{t+1}$ is removed, respectively. The results show that input with $x_t$ and $x_{t+1}$ simultaneously is necessary to estimate fine stiffness information as discussed in Section \ref{sec:method}. 
Moreover, $F(x_{t+1}-x_{t})$ means that we only use the adaptor in StepNet as the self-attention module following the normal paradigm without considering the better estimation of stiffness information in Eq.(\ref{eq:ours}). Comparing the result of $F(x_{t+1}-x_{t})$ and our $F(x_{t+1}, x_t)$, we can see that a finer estimation of stiffness information like ours is necessary to achieve better performance. For (2), as shown in Fig.\ref{fig:ablation}, we constructed four alternative adaptor structures. The experimental results in Table \ref{tab:ablation} illustrate that all four alternatives are inferior to ours. However, the best structure to generate the adaptive step sizes is still unknown, and in the future, we can still improve the design of the adaptor, \textit{e.g.,} to better utilize $x_t$ and $x_{t+1}$ through a neural network effectively.

%在2.2节中我们总结了注意力机制从两个方面，即提取好的刚性信息以及合适的步长的产生，来帮助模型性能的提升。而这两点对应着注意力模块的输入和模块的神经网络结构设计。(1) 对于注意力模块的输入，在表3中，我们探究图4b所示的adaptor在xt 或者xt+1分别被去除之后的模型性能。实验表明xt和xt+1的同时输入是必要的，是估计好的stiffness information的有效输入。(2) 对于模块的神经网络结构，如图5所示，我们构造了4种额外的adaptor。表3的实验结果说明，目前如图4b所示的结构可以获得最佳的性能。但这不意味着它是产生合适注意力值得最佳结构，在未来，adaptor仍然需要更多的改进。
\begin{table}[t]
\small
  \centering
  % \vspace{-0.15cm}
  \resizebox*{1.0\linewidth}{!}{
    \begin{tabular}{lccc}
    \toprule
    \textbf{Model} & \multicolumn{1}{l}{\textbf{CIFAR10}} & \multicolumn{1}{l}{\textbf{CIFAR100}} & \multicolumn{1}{l}{\textbf{STL10}} \\
    \midrule
    % Only $x_t$    &  94.59$\pm$ 0.14     &  76.34$\pm$ 0.14     & 84.74$\pm$ 0.64 \\
    % Only $x_{t+1}$  &  94.32$\pm$ 0.23     &   76.53$\pm$ 0.04    & 85.12$\pm$ 0.54 \\
    Org  & 93.35$\pm$ 0.18            & 74.30$\pm$ 0.30            &  82.66$\pm$ 1.05      \\
    $F(x_t)$  & 94.59$\pm$ 0.14            & 76.34$\pm$ 0.14            &  84.74$\pm$ 0.64      \\
    $F(x_{t+1})$ &  94.32$\pm$ 0.23           &  76.53$\pm$ 0.04           &  85.12$\pm$ 0.54      \\
    $F(x_{t+1} - x_t)$   & 94.69$\pm$ 0.33             &  76.42$\pm$ 0.10           &  85.26$\pm$ 0.72      \\
    $F(x_{t+1}, x_t)$ (ours) & \textbf{95.14}$\pm$ \textbf{0.24}            &  \textbf{77.04}$\pm$\textbf{0.22}          &  \textbf{86.08}$\pm$\textbf{0.11}      \\
    \midrule
    Adaptor \textbf{a}    &  94.39$\pm$ 0.12     &  76.19$\pm$ 0.14     & 84.21$\pm$ 0.12 \\
    Adaptor \textbf{b}  &  94.05$\pm$ 0.11     &  75.45$\pm$ 0.12     &  84.10$\pm$ 0.24\\
    Adaptor \textbf{c}  &  94.19$\pm$ 0.24     &  75.12$\pm$ 0.08     &  84.12$\pm$ 0.44\\
    Adaptor \textbf{d}  &  94.92$\pm$ 0.24     &   76.83$\pm$ 0.14    & 85.62$\pm$ 0.44 \\
    \bottomrule
    \end{tabular}%
    }
    \caption{The ablation study on the structure of StepNet.}
    \vspace{-15pt}
  \label{tab:ablation}%
\end{table}%

\textbf{The property of stiffness phenomenon (SP) and the LTH4SA.}
In fact, for any input $x_0$ and its corresponding output $y_0$, the feature trajectory built by a well-trained residual neural network has two properties: (1) For most inputs $x_0$, their feature trajectories have SP; (2) For each trajectory, only a few features can cause SP. Specifically, for property (1), in Section \ref{sec:gtt}, we approximate the GT trajectories with several advanced self-attention models. From Fig.\ref{fig:totalstiff} and Fig.\ref{fig:stiffness_vis}, we can empirically observe that for most of the inputs, their feature trajectories have SP.
In the appendix, we provide more visualizations of feature trajectories using SENet as an example, and these visualizations also provide empirical evidence for property (1). For property (2), we define the stiffness proportion $\hat{p} = \frac{1}{L}\mathbb{E}_{x_0\sim P(x_0)} \#\{ t| \zeta_{\text{NSI}}(x_t) \geq \max(\mu(1+M_1 ),M_2)\}$ to measure the expected number that how many features from a feature trajectory have $\zeta_{\text{NSI}}(x_t;\mathbf{M})$ with degree $\mathbf{M}$.
If the $\hat{p}$ of a feature trajectory is close to 100\%, it means that this trajectory has many features with large enough NSI. For Fig.\ref{fig:arch}, we show the corresponding stiffness proportion in Fig.\ref{fig:totalratio}. We can observe that for various $M_1$ and $M_2$, most of the $\hat{p}\leq $ 10\%, which indicates only a few features in GT trajectory can cause SP. 

Moreover, these two properties are also consistent with those intrinsic properties in physics dynamical systems. For instance, the close encounter is an important factor to cause the SP in the three-body motion. \cite{liang2020instance} considers 1,000 independent simulations of three-body trajectories following \cite{chen2019symplectic}, where 91.4\% of the trajectories contain the close encounter, but on average only $4.2\%$ of the time intervals within the trajectories contain close encounter. 

%Such an analogy may implies that the two properties we identified may also be the intrinsic properties for residual neural networks.
% Therefore, these two properties may be the intrinsic properties of residual neural networks.

In addition, we identify that these two properties are closely related to the lottery ticket hypothesis \cite{huang2022lottery} in the self-attention mechanism (LTH4SA). LTH4SA reveals that we only need to insert the self-attention module on a small number of blocks to achieve remarkable improvement for a NN. According to property (1), for most inputs, their GT trajectories have SP, and as mentioned in Section \ref{sec:self_and_step}, the adaptive step size generated by the SAM can improve the representational ability of NN. Thus the SAM is valid for most of the inputs. Moreover, property (2) tells us that only a small part of the features in a feature trajectory can cause SP, and thus we only need to set the module on a small number of blocks to measure the SP of the whole trajectory. So if these two properties generally hold, we argue that LTH4SA may also be an intrinsic property of the SAM.

\begin{figure}[t]
\centering
\includegraphics[width=1.0\linewidth]{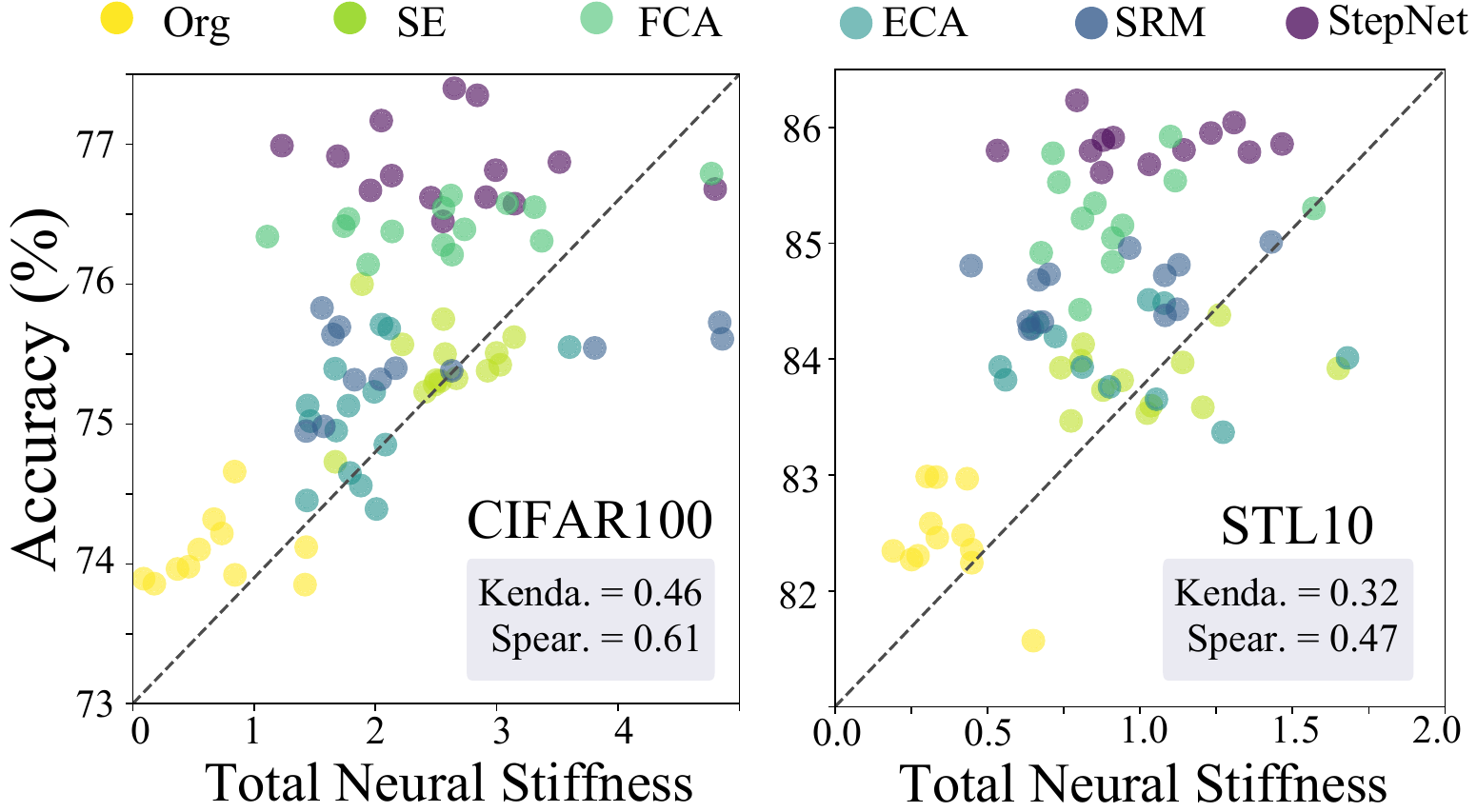}
\vspace{-10pt}
\caption{The correlation between the accuracy and TNS.} 
\label{fig:corr_acc_tns}
\vspace{-15pt}
%尽管小但是也有一些特征 也需要这种特征 允许产生注意力，highlight
\end{figure}

\textbf{Why do the GT trajectories have SP?  } 
Now we attempt to understand why most GT trajectories have SP, which can help us design novel methods to boost the performance of representation learning. From Eq.(\ref{eq:sai_neural}) and Eq.(\ref{eq:stiffbyorg}), we know that for a well-trained residual neural network, $f(x_t;\theta_t)$ provides stiffness information and $\zeta_\text{NSI}(x_t)=\mathcal{O} (\|f(x_t;\theta_t)\|_2)$. When NSI is large, $\|f(x_t;\theta_t)\|_2$ is also large, which means that the elements (absolute values) of the output feature of the neural network $f(\cdot;\theta_t)$ at $t$-th block are relatively large. In some previous works \cite{li2016pruning,hu2016network,polyak2015channel}, such kinds of features are considered important features and have major contributions to the model performance. In other words, a residual network can achieve high performance, \textit{i.e.}, it can approximate the GT trajectory, probably because the network has the ability to learn such important (stiff) features by the adaptive step sizes in a few blocks. So we further calculate the rank correlation (kendall correlation \cite{kendall1938new} and spearman correlation \cite{myers2013research}) between the TNS and the model performance. The results are presented in Fig.\ref{fig:corr_acc_tns}, which shows that the performance of the models and their representational ability to measure the SP are positively correlated. Moreover, as the TNS can reflect the ability of the model to measure the SP, thus the TNS can also be a novel representational ability metric to evaluate the neural network in practice and has the potential to be used in network formulation, such as neural architecture search \cite{huang2020efficient,liu2018darts}, network pruning \cite{li2016pruning,he2021cap,he2021blending} or other applications \cite{ChenLCHW22pami,chen2022cross}.

%我们尝试理解为什么gtt是stiff的, which can help us design the novel method to boost the performance for representation learning.由公式4和公式8，我们知道对于一个well-trained的残差网络，f提供了刚性信息，且NSI=$\mathcal{O} (\|f(x_t;\theta_t)\|_2)$. 当nsi大的时候，$\|f(x_t;\theta_t)\|_2$也大，这意味着神经网络f输出的特征的元素比较大。在过去的一些剪枝工作中，这样的特征被认为是重要的特征且对模型的性能有较大的贡献。换句话说，一个残差网络可以取得很高的性能，即逼近gtt，可能是因为这个网络有能力学习到这种重要的特征。进一步地，我们可以通过sp和kenda去计算the feature traj from 高性能残差网络，被认为是gtt的proxies，的TNS和模型性能的秩相关性。结果在图7中展示，可以发现，模型的性能和他们捕获刚性的能力是正相关的。在未来的工作中，我们可以设计更新的刚性度量，并将这样的度量用于模型的结构设计中，如高效神经网络结构设计，NAS或模型剪枝的准则。
%adaptedly step size自适应步长所导致的？要联系起来

\section{Conclusion}
%在本文中，We 将神经网络中的注意力机制和numerical solution of stiff ordinary differential equations联系起来，并提出了a novel understanding of the Self-attention mechanism。我们的观点有助于理解注意力机制是如何增强模型性能的

In this paper, we bridge the relationship between the self-attention mechanism~(SAM) and the numerical solution of stiff ordinary differential equations, which reveals that the SAM is a stiffness-aware step size adaptor that can refine the estimation of stiffness information and generate suitable attention values for adaptively measuring the stiffness phenomenon in the neural network~(NN) to enhance the representational ability of the NN and achieve high performance.

\section{Acknowledgment}
This work was supported in part by  National Key R\&D Program of China under Grant No. 2020AAA0109700, National Natural Science Foundation of China (NSFC) under Grant No.61836012, Guangdong Basic and Applied Basic Research Foundation No. 2023A1515011374, National Natural Science Foundation of China (NSFC) under Grant No. 62206314, GuangDong Basic and Applied Basic Research Foundation under Grant No. 2022A1515011835.
% effectively alleviate the error accumulation in the neural network due to the stiffness issue, thus ensuring the high model performance. 

{\small
\bibliographystyle{ieee_fullname}
\bibliography{egbib}
}

\clearpage

\onecolumn
% \appendix
\section*{Appendix A}
\begin{figure}[h]
	 		\centering
	 		\includegraphics[width=1.0\linewidth]{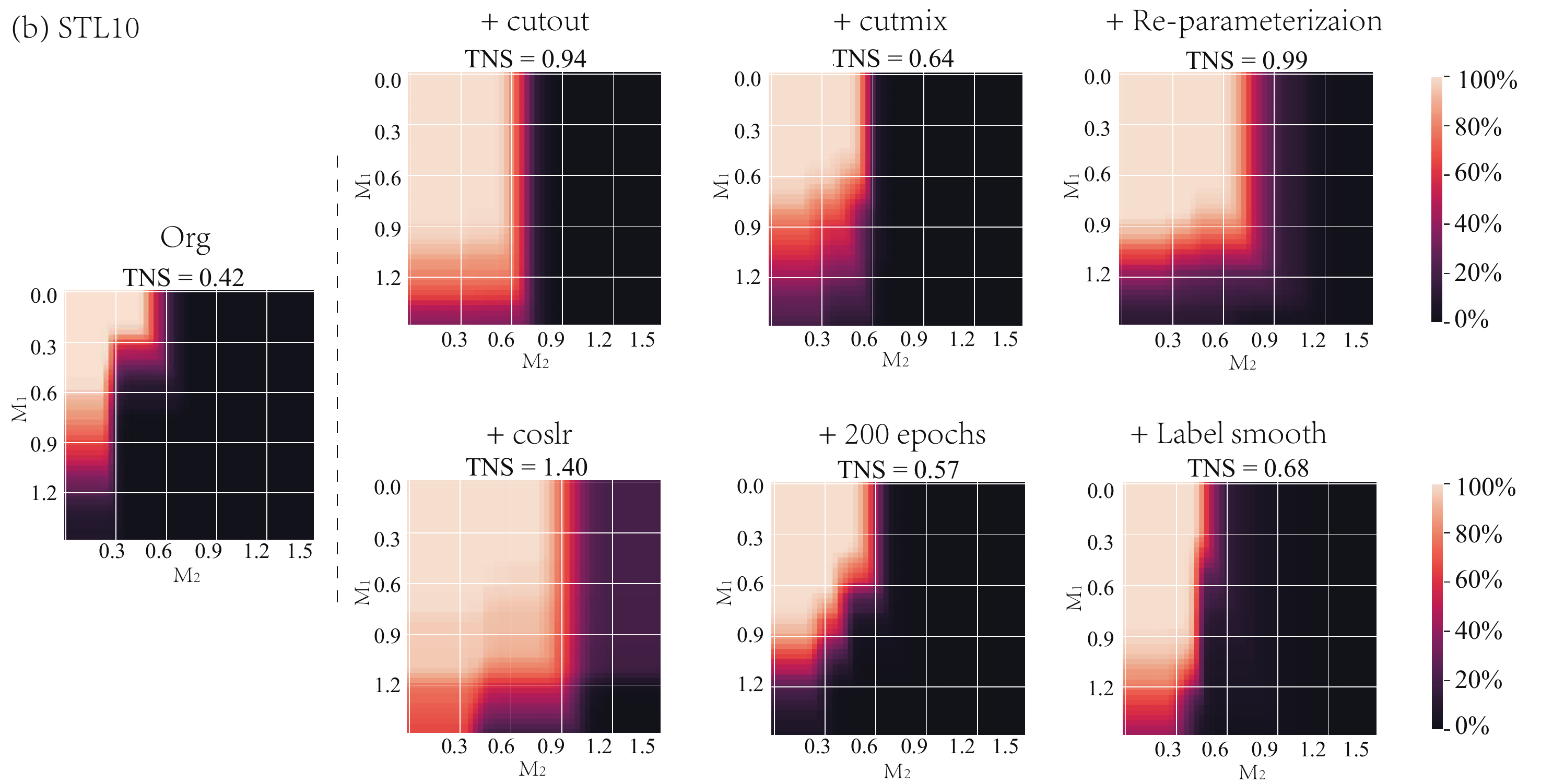}
	 				\vspace{-0.1cm}
	 		\caption{The TNS results for other high-performance training tricks.} 
	 \vspace{-0.2cm}
	 		\label{fig:othermethod}
	 		% \vspace{-0.3cm}
	 	\end{figure}

In addition to the TNS results on the self-attention mechanism given in the main text, we additionally provide the TNS with results for other training methods that can improve the model performance. From Fig.\ref{fig:othermethod}, we again verify that the high-performance models have a strong representational ability to measure the stiffness phenomenon, which is consistent with the results shown in our main text.
	 	 	\begin{figure}[h]
	 		\centering
	 		\includegraphics[width=1.0\linewidth]{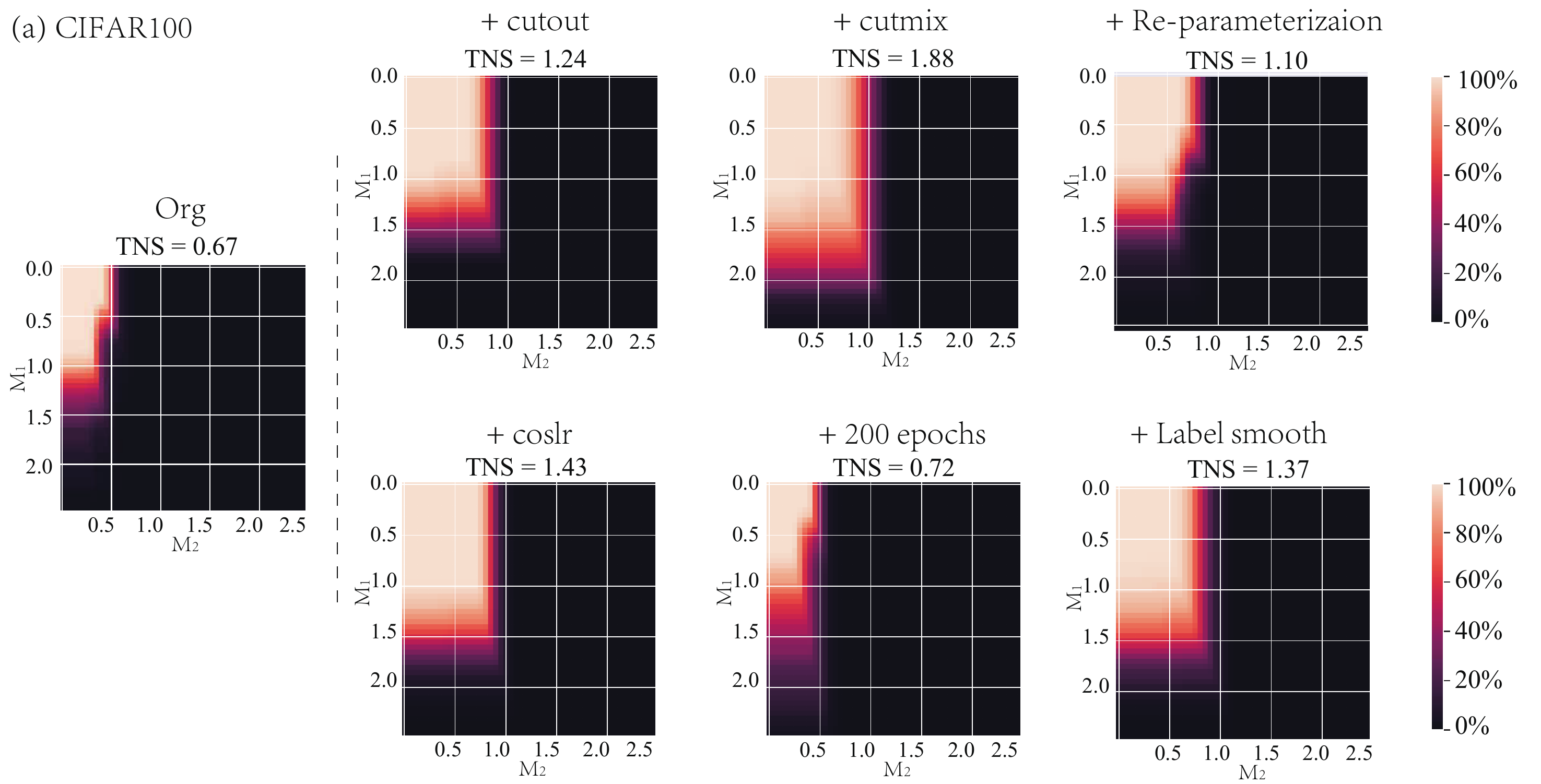}
	 				\vspace{-0.1cm}
	 		% \caption{The TNS results about other .} 
	 \vspace{-0.2cm}
	 		\label{fig:}
	 		% \vspace{-0.3cm}
	 	\end{figure}
   
\clearpage
\section*{Appendix B}
 \textbf{Lemma 1}. Given the feature trajectories $x_{1},x_{2},x_{3}...,x_{L}$ generated by a neural network with $L$ residual blocks, \textit{i.e.,} $x_{t+1} = x_t + f(x_t;\theta_t)\cdot \Delta_t, t=0,1,..,L-1$, where the norm $\|x_t\|_2$ and step size $\Delta t$ are bounded. For $\delta(\mathbf{M})$ defined as $\delta(\mathbf{M}) = \mathbb{E}_{x_0\sim P(x_0)} \mathbf{I}_{\exists t, s.t. \zeta_{\text{NSI}}(x_t) \geq \max(\mu(1+M_1),M_2)}$, $\exists \tilde{M} \in \mathbb{R}+$, s.t. if $\min(M_1,M_2) > \tilde{M}$, $\delta(\mathbf{M})=0$.
\begin{proof}
	Let $0 < k_1 \leq \|x_t\|_2 \leq k_2, t = 1,2,...,L-1$ and $\Delta_t \in [a,b]$, where $k_1,k_2,a$ and $b \in \mathbb{R}_+$. In fact this condition is practical and mild in neural networks. Next, we prove that $\exists \tilde{M} \in \mathbb{R}_+$ s.t. when $M_1 > \tilde{M}$ and $M_2 > \tilde{M}$, we have (1) $\zeta_{\text{NSI}}(x_t) < M_2$ and (2) $\zeta_{\text{NSI}}(x_t) < \mu(1+M_1)$.
	For (1) $\zeta_{\text{NSI}}(x_t) < M_2$, since the boundary and triangles inequality,

	\begin{equation}
	\begin{aligned}
	\zeta_{\text{NSI}}(x_t)&= \frac{1}{\|x_t\|_2}\big\|\frac{x_{t+1}-x_t}{\Delta_t}\big\|_2\\ 
	&= \frac{1}{\Delta_t}\frac{1}{\|x_t\|_2}\|x_{t+1}-x_t\|_2\\
	&\leq \frac{1}{a}\cdot\frac{1}{\|x_t\|_2}\cdot (\|x_{t+1}\|_2 + \|x_{t}\|_2) = \frac{1}{a}(1 + \frac{\|x_{t+1}\|_2}{\|x_{t}\|_2}) \leq \frac{1}{a}(1 + \frac{k_2}{k_1})  
	\end{aligned}
	\label{eq:upper_nsi}
	\end{equation}
	Therefore, if $M_2 > \frac{1}{a}(1 + \frac{k_2}{k_1})$, we have $\zeta_{\text{NSI}}(x_t) < M_2$.
	For (2) $\zeta_{\text{NSI}}(x_t) < \mu(1+M_1)$, we can first estimate the lower bound of $\zeta_{\text{NSI}}(x_t)$.
	
	\begin{equation}
	\begin{aligned}
	\zeta_{\text{NSI}}(x_t)&= \frac{1}{\|x_t\|_2}\big\|\frac{x_{t+1}-x_t}{\Delta_t}\big\|_2\\ 
	&= \frac{1}{\Delta_t}\frac{1}{\|x_t\|_2}\|x_{t+1}-x_t\|_2\\
	&\geq \frac{1}{b}\cdot\frac{1}{\|x_t\|_2}\cdot |\|x_{t+1}\|_2 - \|x_{t}\|_2| = \frac{1}{b}\cdot \frac{1}{k_2} \cdot |k_1 - k_2|. 
	\end{aligned}
	\label{eq:lower_nsi}
	\end{equation}
	
	Moreover, note that 
	
	\begin{equation}
	\begin{aligned}
	\zeta_{\text{NSI}}(x_t) < \mu(1+M_1)&\Leftrightarrow \zeta_{\text{NSI}}(x_t) < \frac{1}{L} \sum_{i=1}^L \zeta_{\text{NSI}}(x_i) (1+M_1)\\ 
	&\Leftrightarrow L\cdot \zeta_{\text{NSI}}(x_t)/\sum_{i=1}^L \zeta_{\text{NSI}}(x_i) - 1 < M_1
	\end{aligned}
	\label{eq:mu_bound}
	\end{equation}
	
	From Eq.(\ref{eq:upper_nsi}), we have $L\cdot \zeta_{\text{NSI}}(x_t) \leq \frac{L}{a}(1 + \frac{k_2}{k_1})$. Furthermore, from Eq.(\ref{eq:lower_nsi}), we have 
	\begin{equation}
	    \frac{1}{\sum_{i=1}^L \zeta_{\text{NSI}}(x_i)} \leq \frac{1}{\sum_{i=1}^L \frac{1}{b}\cdot \frac{1}{k_2} \cdot |k_1 - k_2|} = \frac{bk_2}{L|k_1-k_2|}.
	    \label{eq:temp1}
	\end{equation}
	
	Therefore, when $M_1$ meets 
	\begin{equation}
	    M_1 > \frac{L}{a}(1 + \frac{k_2}{k_1}) \cdot \frac{bk_2}{L|k_1-k_2|} - 1 = \frac{bk_2(k_1+k_2)}{ak_1\cdot |k_1-k_2|} - 1,
	\end{equation}
	
	we have $M_1 > L\cdot \zeta_{\text{NSI}}(x_t)/\sum_{i=1}^L \zeta_{\text{NSI}}(x_i) - 1$ and $\zeta_{\text{NSI}}(x_t) < \mu(1+M_1)$ holds. Let 
	\begin{equation}
	    \tilde{M} = \max\big(\frac{bk_2(k_1+k_2)}{ak_1\cdot |k_1-k_2|} - 1,\frac{1}{a}(1 + \frac{k_2}{k_1})\big).
	\end{equation}

	When $\min(M_1,M_2) > \tilde{M}$, for any $t$,
	\begin{equation}
	     \zeta_{\text{NSI}}(x_t) < \max( M_2 , \mu(1+M_1)),
	\end{equation}
	
	Therefore,
	\begin{equation}
	    \delta(\mathbf{M}) = \mathbb{E}_{x_0\sim P(x_0)} \underbrace{\mathbf{I}_{\exists t, s.t. \zeta_{\text{NSI}}(x_t) \geq \max(\mu(1+M_1),M_2)}}_{\text{equal to 0}} = 0.
	\end{equation}

	\qedhere
\end{proof}

% \clearpage

% \section*{Appendix C}
\textbf{Theorem 1}.
 For the $\delta(\mathbf{M})$ defined as $\delta(\mathbf{M}) = \mathbb{E}_{x_0\sim P(x_0)} \mathbf{I}_{\exists t, s.t. \zeta_{\text{NSI}}(x_t) \geq \max(\mu(1+M_1),M_2)}$, the Total Neural Stiffness (TNS) $\iint_{\mathbf{M}} \delta(\mathbf{M}) d\mathbf{M}$ is convergent.

\begin{proof}
 Let 
	\begin{equation}
	    \tilde{M} = \max\big(\frac{bk_2(k_1+k_2)}{ak_1\cdot |k_1-k_2|} - 1,\frac{1}{a}(1 + \frac{k_2}{k_1})\big).
	\end{equation}
	
		Note that $0\leq M_1,M_2$,

	\begin{equation}
	\begin{aligned}
	\iint_{\mathbf{M}} \delta(\mathbf{M}) d\mathbf{M} &= \int_0^{+\infty}\int_0^{+\infty} \delta(M_1,M_2) dM_1 dM_2\\ 
	&=\int_0^{\tilde{M}}\int_0^{\tilde{M}} \delta(M_1,M_2) dM_1 dM_2 + \underbrace{\int_{\tilde{M}}^{+\infty}\int_{\tilde{M}}^{+\infty} \delta(M_1,M_2) dM_1 dM_2}_{\text{equal to 0 since Lemma 1}}.\\
	\end{aligned}
	\label{eq:temp11}
	\end{equation}

	Since $0 \leq \mathbf{I}_{\exists t, s.t. \zeta_{\text{NSI}}(x_t) \geq \max(\mu(1+M_1),M_2)} \leq 1$, we have 
	\begin{equation}
	    0\leq \delta(\mathbf{M}) = \mathbb{E}_{x_0\sim P(x_0)} \mathbf{I}_{\exists t, s.t. \zeta_{\text{NSI}}(x_t) \geq \max(\mu(1+M_1),M_2)} \leq 1.
	    \label{eq:temp12}
	\end{equation}
	
	Therefore,
	
% 	\begin{equation}
	\begin{align*}
	\iint_{\mathbf{M}} \delta(\mathbf{M}) d\mathbf{M} 
	&=\int_0^{\tilde{M}}\int_0^{\tilde{M}} \delta(M_1,M_2) dM_1 dM_2 + 0 \tag*{from Eq. (\ref{eq:temp11})}\\
	&\leq \int_0^{\tilde{M}}\int_0^{\tilde{M}}  dM_1 dM_2 < +\infty \tag*{from Eq. (\ref{eq:temp12})}
	\end{align*}
% 	\label{eq:temp23}
% 	\end{equation}	
	
	The $\delta(\mathbf{M})$ is positive and $\iint_{\mathbf{M}} \delta(\mathbf{M}) d\mathbf{M}$ is bounded, thus the Total Neural Stiffness $\iint_{\mathbf{M}} \delta(\mathbf{M}) d\mathbf{M}$ is convergent.

	\qedhere
\end{proof}

\clearpage

\section*{Appendix C}

 \textbf{Theorem 2}. For an ordinary differential equation $\text{d}\mathbf{u}/\text{d}t = \mathbf{f}(\mathbf{u})$, if the Jacobian matrix $\mathbf{J}_{u^t}$ at $u^t$ is a $n\times n$ symmetric real matrix and $\{\lambda_i\}_{i=1}^n$ are its $n$ distinct eigenvalues, and $\mathbf{Re}(\lambda_i)<0,i=1,2,...,n$, then
 \begin{equation}
     \zeta_{\text{SAI}}(u^{t})  \approx   \zeta_{\text{SI}}(u^t) \cdot \sqrt{c + Q[\zeta_{\text{SI}}(u^t)]},
 \end{equation}
 where $c$ is a constant and $Q(\cdot)$ is a function with respect to $\zeta_{\text{SI}}(u^t)$ and when $\zeta_{\text{SI}}(u^t)$ is large enough, $Q[\zeta_{\text{SI}}(u^t)]$ converges to a 0.

\begin{proof}
	
	Note that $\zeta_{\text{SAI}}(\cdot)$ is computed by adjacent states with small step size, and the adjacent states are closed to a linearized ODE. Therefore, we use Taylor expansion to provide a reasonable approximation for the right-hand side of the equation. Specifically, we consider the Taylor expansion at $u^t$ for $\mathbf{f}(\mathbf{u})$, we have

	\begin{equation}
	\begin{aligned}
	\mathbf{f}(\mathbf{u}) &=  \mathbf{f}(u^t) + \mathbf{J}_{u^t} (\mathbf{u} - u^t) + o(\|\mathbf{u} - u^t\|)\\
	&= \mathbf{J}_{u^t}\mathbf{u} +\mathbf{f}(u^t) - \mathbf{J}_{u^t}u^t + o(\|\mathbf{u} - u^t\|)\approx \mathbf{J}_{u^t}\mathbf{u} + \mathbf{h}(t).\\
	\end{aligned}
	\label{eq:temp323}
	\end{equation}	
	
	Let $\{\mathbf{v}_i\}_{i=1}^n$ be the eigenvectors corresponding to the eigenvalues $\{\lambda_i\}_{i=1}^n$. Since the Jacobian matrix $\mathbf{J}_{u^t}$ at $u^t$ is a $n\times n$ symmetric real matrix, $\{\mathbf{v}_i\}_{i=1}^n$ form a set of orthogonal vectors.  Without loss of generalization, we assume $\{\mathbf{v}_i\}_{i=1}^n$ is standard orthogonal basis. Therefore, $u^t$ can be represented by the basis $\{\mathbf{v}_i\}_{i=1}^n$. Let
	\begin{equation}
	    \mathbf{u}(0) = u^t = \sum_{i=1}^n c_i\mathbf{v}_i,
	    \label{eq:temppsf}
	\end{equation}
	where $c_i \in \mathbb{R}$. Moreover, Eq.(\ref{eq:temp323}) is a linear constant coefficient inhomogeneous equation. The solution of this equation is
	\begin{equation}
	    \mathbf{u}(t)=\sum_{i=1}^nc_i\mathbf{v}_i e^{\lambda_it} + \mathbf{g}(t),
	\end{equation}
	 where $\mathbf{g}(t)$ is steady-state solution and since Eq.(\ref{eq:temppsf}), $\mathbf{g}(0) = \mathbf{0}$. Let $u^{t^{\prime}} = \mathbf{u}(\Delta t)$, we have
	
	\begin{align*}
	\lim_{\Delta t \to 0} \frac{1}{\|u^{t}\|_2}\big\|\frac{u^{t^{\prime}}-u^{t}}{\Delta t}\big\|_2.
	&=\lim_{\Delta t \to 0}\frac{1}{\|u^{t}\|_2}\big\|\frac{\sum_{i=1}^nc_i\mathbf{v}_i e^{\lambda_i\Delta t} + \mathbf{g}(\Delta t) - \sum_{i=1}^n c_i\mathbf{v}_i}{\Delta t}\big\|_2 \\
	&= \lim_{\Delta t \to 0}\frac{1}{\|u^{t}\|_2}\big\|\frac{\sum_{i=1}^nc_i\mathbf{v}_i (e^{\lambda_i\Delta t}-1) }{\Delta t} + \frac{ \mathbf{g}(\Delta t) - \mathbf{g}(0)}{\Delta t}  \big\|_2\\
	&= \frac{1}{\|u^{t}\|_2} \big\|\sum_{i=1}^n\lambda_i c_i\mathbf{v}_i +  \nabla\mathbf{g}(0)  \big\|_2 \tag*{Since $e^x-1\sim x$}\\
	\end{align*}	
	%h(t)没有也可以得到类似的结论
	
	And $\nabla\mathbf{g}(0)$ can be linear combination by the standard orthogonal basis $\{\mathbf{v}_i\}_{i=1}^n$. Let $\nabla\mathbf{g}(0) = \sum_{i=1}^n a_i\mathbf{v}_i$, where $a_i \in R$. Note that 
	\begin{equation}
	    \|u^{t}\|_2 = \|\sum_{i=1}^nc_i\mathbf{v}_i\|_2 = (\sum_{i=1}^n c_i^2)^{1/2},
	    \label{eq:temopppd}
	\end{equation}
	we can find that
	
	\begin{align*}
	\lim_{\Delta t \to 0} \frac{1}{\|u^{t}\|_2}\big\|\frac{u^{t^{\prime}}-u^{t}}{\Delta t}\big\|_2.
	&= \frac{1}{\|u^{t}\|_2} \big\|\sum_{i=1}^n\lambda_i c_i\mathbf{v}_i +  \nabla\mathbf{g}(0)  \big\|_2\\
	&= \frac{1}{\|u^{t}\|_2} \big\|\sum_{i=1}^n (\lambda_i c_i + a_i)\mathbf{v}_i   \big\|_2 \tag*{Since $\nabla\mathbf{g}(0) = \sum_{i=1}^n a_i\mathbf{v}_i$}\\
	&= \big[\frac{\sum_{i=1}^n(\lambda_ic_i+a_i)^2}{\sum_{i=1}^nc_i^2}\big]^{\frac{1}{2}}.\tag*{Since Eq.(\ref{eq:temopppd})}
	\end{align*}
	
% 	Note that $\zeta_{\text{SI}}(u^t) = \max(|\mathbf{Re}(\lambda_i)|)$, w
	Without loss of generalization, we assume  $|\mathbf{Re}(\lambda_1)|\geq |\mathbf{Re}(\lambda_2)| \geq ... \geq |\mathbf{Re}(\lambda_n)|$. Moreover, since the matrix $J_{u^t}$ is symmetric real matrix, the eigenvalues are real number, \textit{i.e.}, $\mathbf{Re}(\lambda_i) = \lambda_i, i =1,2,...,n$. Therefore, 
	\begin{equation}
	    \zeta_{\text{SI}}(u^t) = \max(|\mathbf{Re}(\lambda_i)|) = |\lambda_1|.
	    \label{eq:sidfsdfs}
	\end{equation}
	
	Moreover,
	\begin{align*}
	\lim_{\Delta t \to 0} \frac{1}{\|u^{t}\|_2}\big\|\frac{u^{t^{\prime}}-u^{t}}{\Delta t}\big\|_2
	&= \big[\frac{\sum_{i=1}^n(\lambda_ic_i+a_i)^2}{\sum_{i=1}^nc_i^2}\big]^{\frac{1}{2}}.\\
	&= \big[  \sum_{i=1}^n   (\frac{c_i^2}{\sum_{i=1}^nc_i^2}\lambda_i^2 + \frac{2a_i}{\sum_{i=1}^nc_i^2}\lambda_i + \frac{a_i^2}{\sum_{i=1}^nc_i^2})   \big]^{\frac{1}{2}}\\
	&= \big[  \frac{c_1^2}{\sum_{i=1}^nc_i^2}\lambda_1^2 + \sum_{i=2}^n   \frac{c_i^2}{\sum_{i=1}^nc_i^2}\lambda_i^2  +\sum_{i=1}^n (\frac{2a_i}{\sum_{i=1}^nc_i^2}\lambda_i + \frac{a_i^2}{\sum_{i=1}^nc_i^2})               \big]^{\frac{1}{2}}\\
	&= \zeta_{\text{SI}}(u^t) \big[  \underbrace{\frac{c_1^2}{\sum_{i=1}^nc_i^2}}_{  \text{Constant}  } + \underbrace{\sum_{i=2}^n   \frac{c_i^2}{\sum_{i=1}^nc_i^2}\frac{\lambda_i^2}{\lambda_1^2}  +\sum_{i=1}^n (\frac{2a_i}{\sum_{i=1}^nc_i^2}\frac{\lambda_i}{\lambda_1^2} + \frac{a_i^2}{\lambda_1^2\sum_{i=1}^nc_i^2})               }_{\text{The term with respect to SI}}\big]^{\frac{1}{2}} \tag*{Since Eq.(\ref{eq:sidfsdfs})}\\
	&\triangleq \zeta_{\text{SI}}(u^t) \cdot \sqrt{(c + Q[\zeta_{\text{SI}}(u^t)])}
	\end{align*}
	
	%近似
	Thus, $\zeta_{\text{SAI}}(u^{t})= \zeta_{\text{SI}}(u^t) \cdot \sqrt{c + Q[\zeta_{\text{SI}}(u^t)]} + \mathcal{O}(\Delta t)$ and according to Eq.(\ref{eq:temp323}), $\zeta_{\text{SAI}}(u^{t}) \approx \zeta_{\text{SI}}(u^t) \cdot \sqrt{(c + Q[\zeta_{\text{SI}}(u^t)])}$ for ordinary differential equation $\text{d}\mathbf{u}/\text{d}t = \mathbf{f}(\mathbf{u})$ holds. Next, since Eq.(\ref{eq:sidfsdfs}),
	
	\begin{equation}
	    \lim_{\zeta_{\text{SI}}(u^t)\to +\infty} Q[\zeta_{\text{SI}}(u^t)] = \lim_{|\lambda_1|\to +\infty} \sum_{i=2}^n   \frac{c_i^2}{\sum_{i=1}^nc_i^2}\frac{\lambda_i^2}{\lambda_1^2}  +\sum_{i=1}^n (\frac{2a_i}{\sum_{i=1}^nc_i^2}\frac{\lambda_i}{\lambda_1^2} + \frac{a_i^2}{\lambda_1^2\sum_{i=1}^nc_i^2}) = 0.
	\end{equation}
	
	In Eq.(\ref{eq:temp323}), if we consider another linear approximation like $\mathbf{f}(\mathbf{u})\approx \mathbf{J}_{u^t}\mathbf{u}$, we can also obtain the similar conclusion as  $\zeta_{\text{SAI}}(u^{t}) \approx \zeta_{\text{SI}}(u^t) \cdot \sqrt{c + U[\zeta_{\text{SI}}(u^t)]} $, where $U(\cdot)$ is a function with respect to $\zeta_{\text{SI}}(u^t)$ and when $\zeta_{\text{SI}}(u^t)$ is large enough, $U[\zeta_{\text{SI}}(u^t)]$ converges to a 0. Specifically,

 	\begin{align*}
	\lim_{\Delta t \to 0} \frac{1}{\|u^{t}\|_2}\big\|\frac{u^{t^{\prime}}-u^{t}}{\Delta t}\big\|_2
    &= \frac{1}{\|u^{t}\|_2} \big\|\sum_{i=1}^n\lambda_i c_i\mathbf{v}_i   \big\|_2\\
	&= \big[\sum_{i=1}^n\frac{c_i^2}{\sum_{i=1}^nc_i^2}\lambda_i\big]^{\frac{1}{2}}\\
	&= \big[  \frac{c_1^2}{\sum_{i=1}^nc_i^2}\lambda_1^2 + \sum_{i=2}^n   \frac{c_i^2}{\sum_{i=1}^nc_i^2}\lambda_i^2  ]^{\frac{1}{2}}\\
	&= \zeta_{\text{SI}}(u^t) \big[  \underbrace{\frac{c_1^2}{\sum_{i=1}^nc_i^2}}_{  \text{Constant}  } + \underbrace{\sum_{i=2}^n   \frac{c_i^2}{\sum_{i=1}^nc_i^2}\frac{\lambda_i^2}{\lambda_1^2}                 }_{\text{The term with respect to SI}}\big]^{\frac{1}{2}} \tag*{Since Eq.(\ref{eq:sidfsdfs})}\\
	&\triangleq \zeta_{\text{SI}}(u^t) \cdot \sqrt{(c + U[\zeta_{\text{SI}}(u^t)])}
	\end{align*}

 Moreover,
 	\begin{equation}
	    \lim_{\zeta_{\text{SI}}(u^t)\to +\infty} U[\zeta_{\text{SI}}(u^t)] = \lim_{|\lambda_1|\to +\infty} \sum_{i=2}^n   \frac{c_i^2}{\sum_{i=1}^nc_i^2}\frac{\lambda_i^2}{\lambda_1^2}   = 0.
	\end{equation}

	\qedhere
\end{proof}

\clearpage

\section*{Appendix D:}

\begin{figure*}[h]
	 		\centering
	 		\includegraphics[width=1.0\linewidth]{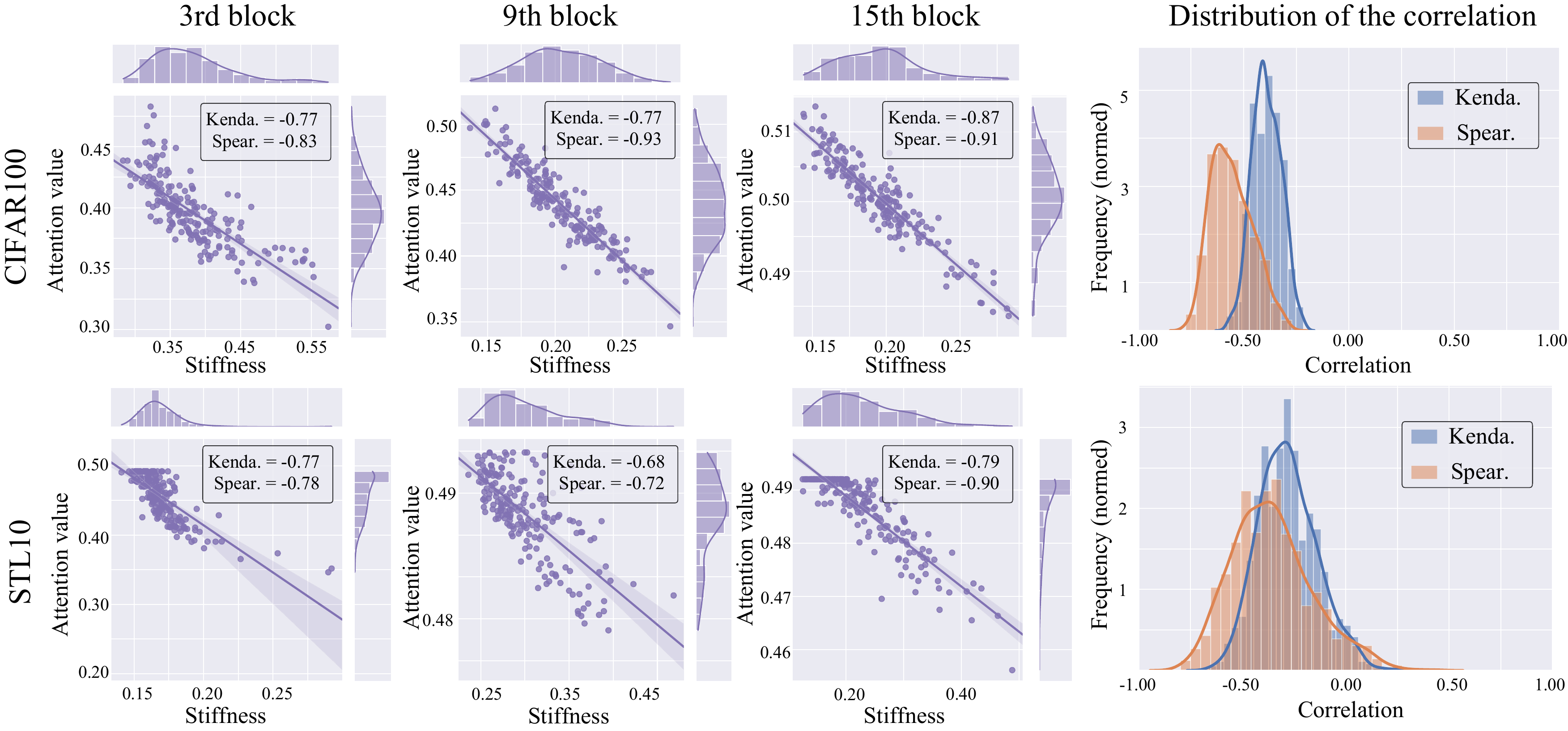}
	 				\vspace{-0.1cm}
	 		\caption{The relationship between the stiffness and attention values (step size). We take the feature trajectories from the first stage of SENet164 as an example. \textbf{a,} the correlations on three specific blocks. \textbf{b,} the distribution of correlations for all trajectories.  } 
	 \vspace{-0.2cm}
	 		\label{fig:case}
	 	\end{figure*}

Since the attention values are generally less than 1 (the attention values are usually measured by the Sigmoid function or Softmax function), they are more fine-grained than the step size $\Delta t = 1$ of the original residual neural networks. 
In various backbones and their different stages, these small step sizes are used in different ways. 
In Fig.\ref{fig:case}, we show an example to understand how the self-attention module uses these small step sizes. Specifically, we take three blocks in the first stage of SENet as an example, we can find that the NSI and attention values are negatively correlated. In other words, in this case, if the stiffness issues exist in the blocks, the module tends to generate a smaller step size to alleviate the stiff issues, which is consistent with the discussions about the stiffness of ODEs in Section 2.1.2.

\clearpage
% gtt不是唯一的，但是性质两个是相都是具备的
\section*{Appendix E:}
	 	 	\begin{figure}[H]
	 		\centering
	 		\includegraphics[width=1.0\linewidth]{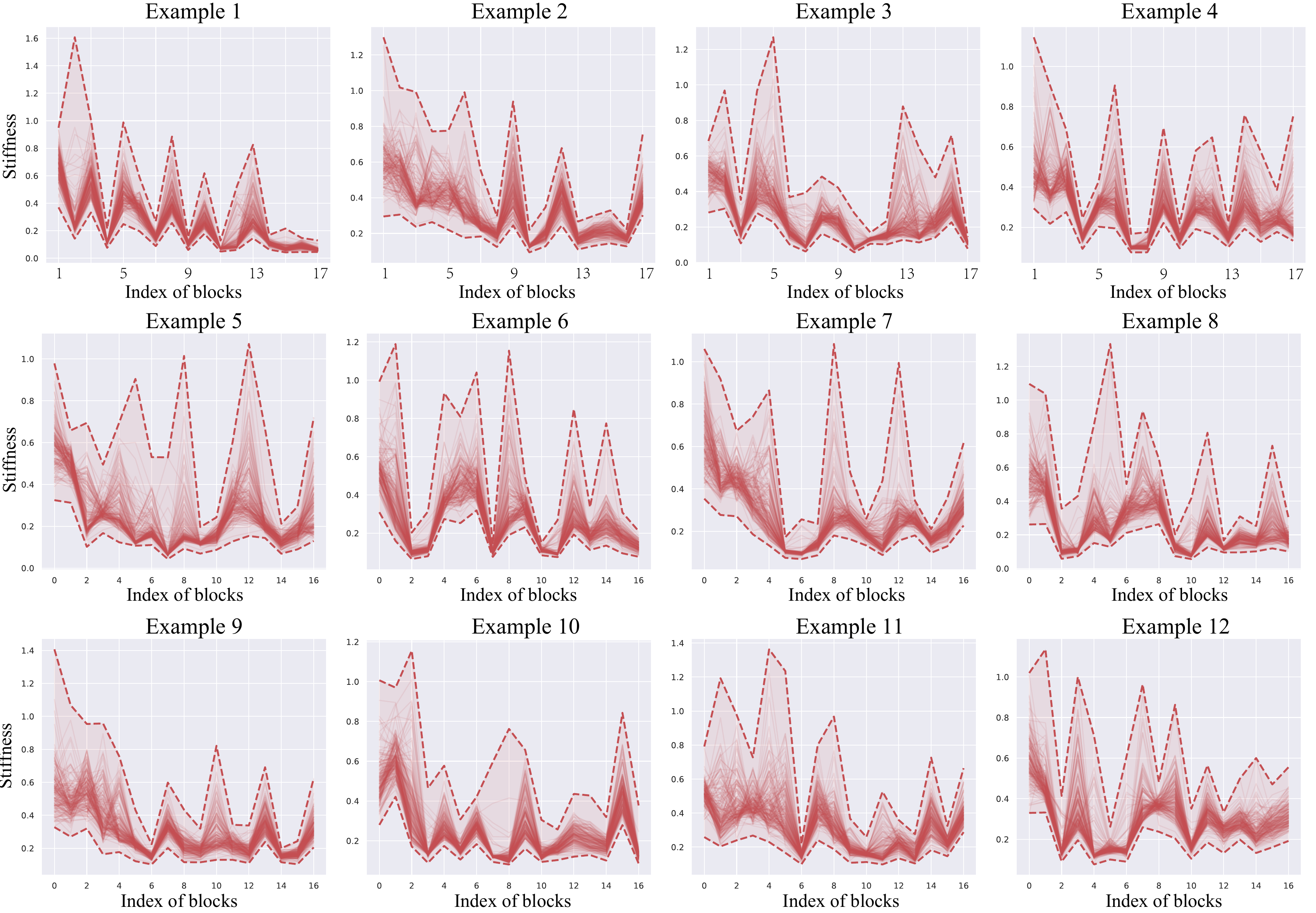}
	 				\vspace{-0.1cm}
	 		\caption{ The visualization of NSI in SENet164 (first stage) on CIFAR100 with different random seeds.} 
	 \vspace{-0.2cm}
	 		\label{fig:dx}
	 		\vspace{-0.3cm}
	 	\end{figure}

\clearpage
	\section*{Appendix F:}
We first introduce the structure of Adaptor in proposed StepNet in main paper. $\sigma$ is \text{Sigmoid} activation function and ``Pooling" is global average pooling. ``Conv" is group convolution with kernel size $k=1$. ``IEBN" \cite{liang2020instance} is the combination of batch normalization and a linear transformation ``IE" from \cite{huang2022layer}. Specifically, for input $x \in R^n$, ``IE" can be written as $\mathbf{W}_{\text{IE}}\otimes x + \mathbf{b}_{\text{IE}}$, where $\mathbf{W}_{\text{IE}} \in R^n$ and $\mathbf{b}_{\text{IE}} \in R^n$ are learnable parameters. The elements in $\mathbf{W}_{\text{IE}}$ and $\mathbf{b}_{\text{IE}}$ can be initialized as 0.0 and -1.0, respectively. All experiments in this paper are verified 5 times with random seeds on RTX 3080 GPUs. We will release our source codes after peer review.

\begin{table}[htbp]
  \centering

    \begin{tabular}{lrrrr}
    \toprule
    \textbf{Dataset} & \multicolumn{1}{l}{\textbf{\#class}} & \multicolumn{1}{l}{\textbf{\#training}} & \multicolumn{1}{l}{\textbf{\#test}} & \textbf{\#Image size} \\
    \midrule
    CIFAR10 & 10    & 50,000 & 10,000 & 32 x 32 \\
    CIFAR100 & 100   & 50,000 & 10,000 & 32 x 32 \\
    STL-10 & 10    & 5,000 & 8,000 & 96 x 96 \\
    % miniImageNet & 100   & 50,000 & 10,000 & 224 x 224 \\
    ImageNet & 1000  & 1,281,123 & 50,000 & 224 x 224 \\
    \bottomrule
    \end{tabular}%
      \caption{The summary of the datasets for image classification experiments.}
  \label{tab:dataclassification}%
\end{table}%

For image classification experiments, the details of the datasets are shown in Table \ref{tab:dataclassification}.
 The hyper-parameter settings of  CIFAR and ImageNet are shown in Table~\ref{tab:imagenet} respectively. For object detection tasks, we consider MS COCO dataset on the same setting as \cite{zhao2021recurrence}. We use Faster R-CNN as detectors, which are implemented by the open-source MMDetection toolkit. The MS COCO dataset contains 80 classes with 118,287 training images and 40,670 test images. “AP”, “AP$_{S}$”,
 “AP$_{M}$”, and “AP$_{L}$” are averaged AP for overall, small, medium, and large scale objects, respectively, at [50\%, 95\%] IoU interval with step as 5\%, “AP$_{50}$” and “AP$_{75}$”: AP at IoU as 50\% and 75\%, respectively.
 
	% \begin{table*}[htbp]
	% 	\small
	% 	\centering
	% 	\begin{tabular}{|c|c|c|c|c|}
	% 		\toprule
	% 		& ResNet164 & PreResNet164 & WRN52-4 & ResNext101-8x32 \\
	% 		\midrule
	% 		Batch size & 128   & 128   & 128   & 128 \\
	% 		Epoch & 180   & 164   & 200   & 300 \\
	% 		Optimizer & SGD(0.9) & SGD(0.9) & SGD(0.9) & SGD(0.9) \\
	% 		depth & 164   & 164   & 52    & 101 \\
	% 		schedule & 60/120 & 81/122 & 80/120/160 & 150/225 \\
	% 		wd    & 1.00E-04 & 1.00E-04 & 5.00E-04 & 5.00E-04 \\
	% 		gamma & 0.1   & 0.1   & 0.2   & 0.1 \\
	% 		widen-factor & -     & -     & 4     & 4 \\
	% 		cardinality & -     & -     & -     & 8 \\
	% 		lr    & 0.1   & 0.1   & 0.1   & 0.1 \\
	% 		$F_{ext}(\cdot)$ & GAP   & BN+GAP & BN+GAP & GAP \\
	% 		drop  & -     & -     & 0.3   & - \\
	% 		\bottomrule
	% 	\end{tabular}%
	% 	\caption{Implementation detail for \textbf{CIFAR10/100} image classification. Normalization and standard data augmentation (random cropping and horizontal flipping) are applied to the training data. GAP and BN denote Global Average Pooling and Batch Normalization separately. }
	% 	\label{tab:cifar}%

	% \end{table*}%

	\begin{table*}[htbp]
		
		\centering
		\begin{tabular}{|c|c|c|c|}
			\toprule
			& ResNet34 & ResNet50 & ResNet164 \\
			\midrule
			Batch size & 256   & 256   & 128    \\
			Epoch & 120   & 120   & 164    \\
			Optimizer & SGD(0.9) & SGD(0.9) & SGD(0.9)  \\
			depth & 34    & 50    & 164    \\
			schedule & 30/60/90 & 30/60/90 & 81/122  \\
			wd    & 1.00E-04 & 1.00E-04 & 1.00E-04  \\
			gamma & 0.1   & 0.1   & 0.1    \\
			lr    & 0.1   & 0.1   & 0.1    \\
			Pooling & GAP   & GAP   & GAP    \\
			\bottomrule
		\end{tabular}%
		\caption{Implementation detail for ImageNet 2012/CIFAR10/CIFAR100 image classification. Normalization and standard data augmentation (random cropping and horizontal flipping) are applied to the training data. For ImageNet dataset, the random cropping of size 224 by 224 is used in corresponding experiments. For CIFAR datasets, we use 32 by 32 size for the random cropping.}

			\label{tab:imagenet}%
	\end{table*}%

\clearpage
\section*{Appendix G:}
\begin{figure}[h]
	 		\centering
	 		\includegraphics[width=0.5\linewidth]{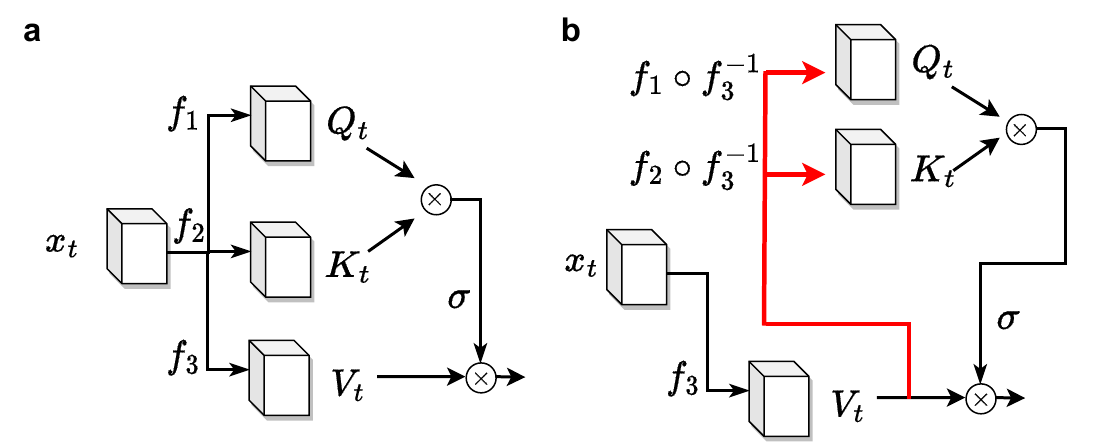}
	 				\vspace{-0.1cm}
	 		\caption{The structure of the transformer-based self-attention mechanism.} 
	 \vspace{-0.2cm}
	 		\label{fig:transformer}
	 		% \vspace{-0.3cm}
	 	\end{figure}

The transformer-based self-attention neural networks are a recently popular residual neural network structure in various artificial intelligence fields. We try to give a preliminary analysis of transformer-based self-attention neural networks by the idea in our main paper. Although there are many variants of this kind of model, we consider the simplest structure shown in Fig.\ref{fig:transformer}a, \textit{i.e},

\begin{equation}
    \hat{x}_{t+1} = x_t + \underbrace{f_3(x_t;\theta_{3t})}_{\text{Feature map}} \otimes \underbrace{\mathbf{F}(f_1(x_t;\theta_{1t}),f_2(x_t;\theta_{2t});\phi_t)}_{\text{Attention value}},
\label{eq:att2}
\end{equation}
where $\sigma$ usually is Softmax activation function, $\phi_t$ is the learnable parameter of the self-attention module, $\mathbf{F}(f_1(x_t;\theta_{1t}),f_2(x_t;\theta_{2t});\phi_t)$ is the attention value. From Eq.(\ref{eq:att2}) and the analysis in main paper, the transformer-based self-attention can also be regarded as the adaptive step size. However, is this kind of step size also stiffness-aware?

In general, $f_3(x_t;\theta_{3t})$,$f_2(x_t;\theta_{2t})$ and $f_1(x_t;\theta_{1t})$ are some learnable matrices. We assume that they are all invertible (they are generally invertible with probability 1 \cite{vershynin2010introduction,huang2020convolution}), as shown in Fig.\ref{fig:transformer}b, and the attention value in Eq.(\ref{eq:att2}) can be rewritten as

	\begin{equation}
	\begin{aligned}
	\mathbf{F}(f_1(x_t;\theta_{1t}),f_2(x_t;\theta_{2t});\phi_t)
	&= \mathbf{F}(f_1\circ f_3^{-1}(V_t)),f_2\circ f_3^{-1}(V_t));\phi_t)\\
	&\triangleq \tilde{\mathbf{F}}(V_t;\tilde{\phi_t}) \\
	&= \tilde{\mathbf{F}}(f_3(x_t;\theta_{3t});\tilde{\phi_t}).
	\end{aligned}
	\label{eq:temp1s1}
	\end{equation}	
Eq.(\ref{eq:temp1s1}) is similar to Eq.(10) in the main paper. At this point, the transformer-based model can also be seen as an adaptive step size adaptor with the stiffness information $f_3(x_t;\theta_{3t}) = \frac{1}{\Delta t}(x_{t+1} - x_t)|_{\Delta t = 1}$ at $x_t$ as input. Of course, this analysis is not necessarily accurate. In fact, the stiffness information at $x_t$ is first about $x_t$. For Eq.(\ref{eq:att2}), it would not be surprising if the stiffness information is provided only by $x_t$. In previous works \cite{guo2020spanet,hu2020squeeze} on channel attention, they also consider $x_t$ as an input to the self-attention module, rather than $f(x_t;\theta_t)$. The experimental results illustrate that the performance of the model can also be improved when $x_t$ is used as input, but the performance of the model with $f(x_t;\theta_t)$ as input is somewhat stronger. This phenomenon may be also attributed to the powerful representational capabilities of neural networks. 

\begin{table}[htbp]
  \centering

  \resizebox{0.5\hsize}{!}{%
    \begin{tabular}{lccc}
    \toprule
    Method & \multicolumn{1}{l}{CIFAR10} & \multicolumn{1}{l}{CIFAR100} & \multicolumn{1}{l}{STL10}\\
    \midrule
    ViT   & $89.08_{(\pm 0.84)}$  & $66.32_{(\pm 1.03)}$ & $62.58_{(\pm 1.02)}$ \\
    ViT+StepNet & $\mathbf{90.32}_{(\pm 0.48)}$      & $\mathbf{68.88}_{(\pm 0.25)}$  &$\mathbf{65.58}_{(\pm 0.59)}$\\
    \bottomrule
    \end{tabular}%
    }
      \caption{The results about ViT with StepNet. All experiments are trained from scratch. }
  \label{tab:vit}%
\end{table}%

% In addition, we do not intend to use StepNet to improve the transformer-based model directly. 

% Because 

Actually, the channel attention and transformer-based models are two views of the self-attention mechanism, the former considers the self-attention mechanism as an additional module that can be plugged into the backbone, and the latter considers the self-attention mechanism as a part of the backbone. 
As shown in Table \ref{tab:vit}, we propose to directly replace the attention modules in ViT with StepNet. Our experimental results demonstrate that the proposed StepNet can indeed be used to enhance the performance of ViT on multiple datasets. Previous works \cite{tolstikhin2021mlp,yu2022metaformer} have also shown that the original ViT can be improved by replacing their attention modules. However, since such a replacement is equivalent to changing the core part of the ViT backbone (transformer-based attention module), can the obtained neural network structures still be called transformer-based methods? This is an issue that deserves further discussion and analysis.

 % As stated in our paper lines 457-458, we have discussed the transformer-based method (TB) \textbf{in Appendix G}. If we use a ViT [15] trained from scratch as the baseline, as shown in the Table below, StepNet can also improve its performance. However, as stated in Appendix G lines 798-800, TB views the attention mechanism as a part of the backbone rather than as an additional module over the Conv layer to model the channel attention. Using StepNet in TB will remove the explicit attention $\text{softmax}(\mathbf{Q}\mathbf{K}^T/\sqrt{d})$ in TB, and ViT-StepNet may not be considered as a TB method. Thus we mainly compare the method belonging to the channel attention method as stated in our paper lines 456-457.
% \section{Introduction}

\end{document}